\documentclass{article}

\usepackage{arxiv}

\usepackage[utf8]{inputenc} 
\usepackage[T1]{fontenc}    
\usepackage{hyperref}       
\usepackage{url}            
\usepackage{doi}                    
\usepackage{booktabs}       
\usepackage{amsfonts}       
\usepackage{nicefrac}       
\usepackage{microtype}      
\usepackage{graphicx}
\usepackage[square,numbers]{natbib}
\usepackage{doi}

\usepackage{amsmath}
\usepackage{amsthm}
\usepackage{cleveref}
\usepackage{algorithm}
\usepackage{algpseudocode}
\usepackage{subfigure}

\usepackage{setspace}
\usepackage{titlesec}
\titlespacing*{\section}{0pt}{*2}{*1}

\usepackage{amssymb}

\newtheorem{lemma}{Lemma}
\theoremstyle{definition}

\theoremstyle{remark}
\newtheorem{remark}{Remark}[section]
 
\newtheorem{proposition}{Proposition}

\DeclareMathOperator{\cov}{cov}
\DeclareMathOperator{\spn}{span}

\title{An Incremental Non-Linear Manifold Approximation Method}

\date{}

\usepackage{authblk}

\author[1]{%
	{\hspace{1mm}Praveen T.~W.~Hettige\thanks{\texttt{hwijewar@mtu.edu}}}
}
\author[1]{%
	{\hspace{1mm}Benjamin W.~Ong \thanks{\texttt{ongbw@mtu.edu}}}
}
\affil[1]{Department of Mathematical Sciences, Michigan Technological University, Michigan, USA}

\hypersetup{
pdftitle={An Incremental Non-Linear Manifold Approximation Method},
pdfsubject={igmra},
pdfauthor={Praveen T.~W.~Hettige,Benjamin W.~Ong},
pdfkeywords={Incremental, Manifold Learning, Multiscale, Non-linear},
}

\begin{document}
\maketitle

\begin{abstract}
Analyzing high-dimensional data presents challenges due to the ``curse of dimensionality'', making computations intensive. Dimension reduction techniques, categorized as linear or non-linear, simplify such data. Non-linear methods are particularly essential for efficiently visualizing and processing complex data structures in interactive and graphical applications.
This research develops an incremental non-linear dimension reduction method using the Geometric Multi-Resolution Analysis (GMRA) framework for streaming data. 
The proposed method enables real-time data analysis and visualization by incrementally updating the cluster map, PCA basis vectors, and wavelet coefficients. Numerical experiments show that the incremental GMRA accurately represents non-linear manifolds even with small initial samples and aligns closely with batch GMRA, demonstrating efficient updates and maintaining the multiscale structure. The findings highlight the potential of Incremental GMRA for real-time visualization and interactive graphics applications that require adaptive high-dimensional data representations.

\end{abstract}


\section{Introduction}
\label{sec1}

High-dimensional data analysis often faces the challenge of the ``curse of dimensionality''. This refers to the difficulty and expense of analyzing data with many parameters and features. Fortunately, high-dimensional data frequently exhibits redundancy or interdependence between parameters and/or features. Dimensionality reduction techniques are valuable tools to capture this redundancy and simplify complex data into a more manageable form. These techniques, broadly categorized as linear and non-linear, are crucial for data visualization, feature extraction.

Linear dimension reduction methods such as Principal Component Analysis (PCA) \cite{pca} efficiently project high-dimensional data onto a hyperplane, a lower-dimensional subspace (a hyperplane). However, in practice, many data sets are distributed along non-linear manifolds.
Non-linear dimension reduction (NLDR) achieves a similar goal as linear dimension reduction methods but projects high-dimensional data into a non-linear lower-dimensional subspace.  NLDR techniques offer valuable capabilities for visualizing complex data sets, reducing the noise in data,
identifying clusters, or as an intermediate step within other analysis
techniques for general data sets. There are several popular NLDR techniques, e.g., Kernel
PCA \cite{scholkopf1997kernel}, Laplacian
Eigenmaps \cite{belkin2003laplacian}, locally-linear
embedding \cite{rowes2000nonlinear}, and
t-SNE \cite{van2008visualizing}, among others. Geometric
Multi-resolution Analysis (GMRA) \cite{ALLARD2012435} is a novel
method of identifying non-linear low-dimensional manifolds and representing data
efficiently. 

Most NLDR techniques were originally designed for static batch data. However, the increasing need for real-time visualization and interactive graphics applications has led to the demand for NLDR methods capable of handling streaming data.

Several NLDR methods have been extended for streaming data dimension reduction. Although GMRA has been applied in streaming data scenarios such as anomaly detection in solar flares \cite{maggioni2013geometric_densityest} and hyperspectral image anomaly detection \cite{chen2012fast_densityest}, no specific algorithm has been proposed to address its application for handling streaming data and addressing potential challenges. This paper proposes a streaming NLDR algorithm based on GMRA for approximating non-linear manifolds. In addition, it explores the challenges associated with using GMRA in the context of streaming data.

\section{GMRA}
\label{sec:gmra}
Geometric multi-resolution analysis (GMRA) \cite{ALLARD2012435} is a somewhat recent method for identifying non-linear low-dimensional manifolds.  The idea is as follows. Suppose that one uses PCA to construct a low-dimensional affine plane to approximate the data. GMRA measures the residual that arises from the affine plane approximation, and if necessary, partitions and computes a correction to the (root-level) affine plane for each partition using the residuals. The combination of the PCA (root-level) affine plane and the (finer-scale) correction gives rise to two-level approximation to the manifold, which can be visualized as a union of affine planes. Iterating this process provides a multi-scale approximation to the manifold. The GMRA framework is particularly attractive because it is data-adaptive, with the ability to generate a non-linear manifold approximation {\em without} over- or under-fitting the data based on residual estimates. Then, a fast transform algorithm is able to map data to the low-dimensional manifold which can be used in applications such as classification or regression. 

\begin{figure}
    \centering
    \includegraphics[width=0.4\textheight,height=0.4\textheight]{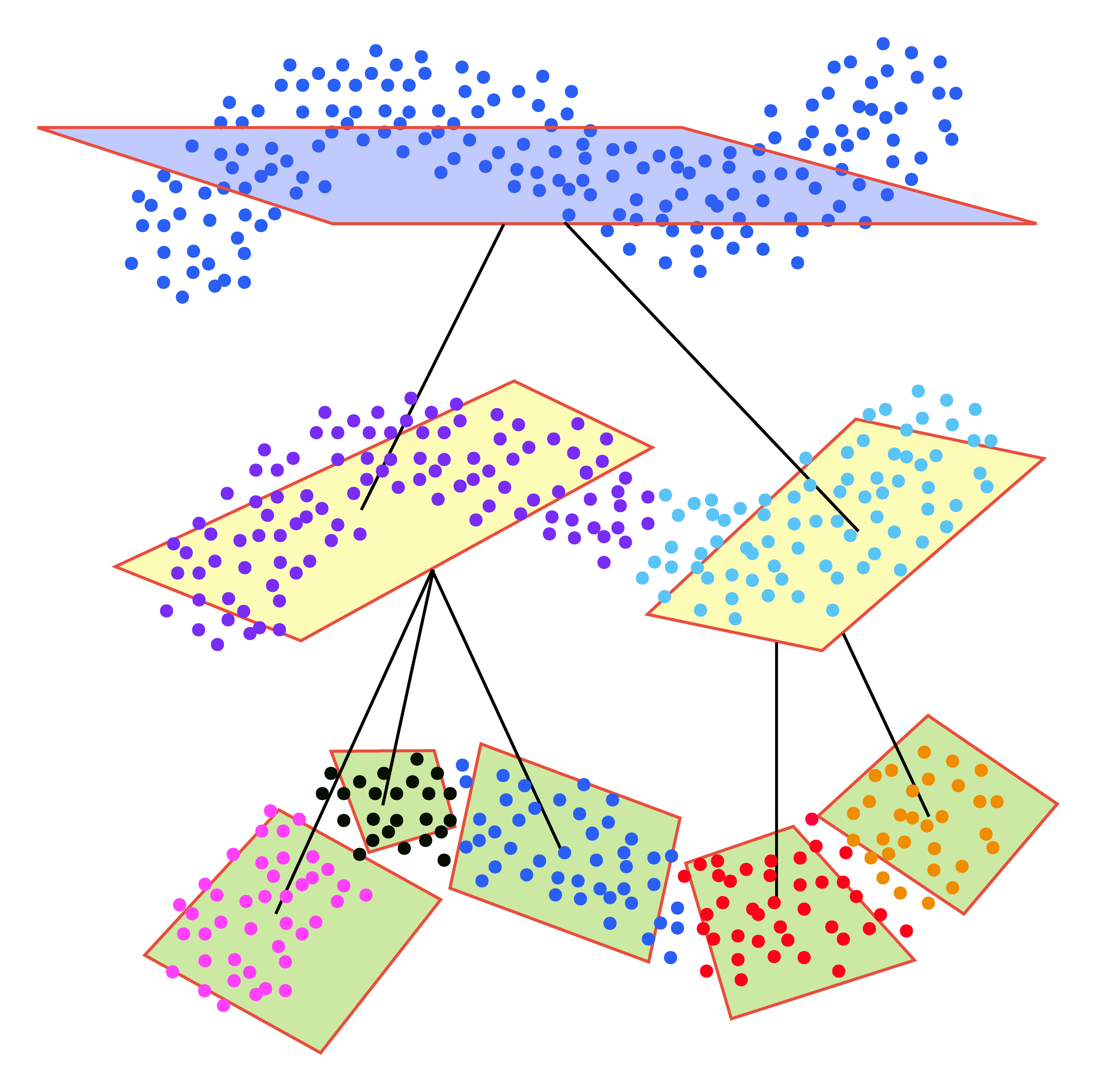}
    \caption{An example illustration of a tree decomposition of point cloud data and hyperplane approximations at each tree node. At the root level, all the sample points are contained within a single node. At the second level, the samples split into two nodes, and the leaf level comprises five tree nodes. Each tree node is represented in a different color.}
    \label{fig:gmra_illustration}
\end{figure}

\Cref{fig:gmra_illustration} shows an illustrative example figure of GMRA construction.
We proceed with a description of the GMRA algorithm and supporting theoretical analysis.

The construction of this GMRA sketch begins with a nested multiscale decomposition of the data $X = \{ \vec{x}_1, \cdots, \vec{x}_N \} \subset \mathbb{R}^D$ into a tree structure.  We utilize a covertree structure \cite{covertree}, which gives rise to partitions that satisfy assumptions for the error estimates. Each tree node corresponds to a nested subset of the data, $C_{s,p} \subseteq X$, that is indexed by both scale $s$, and relative spatial position $p$.  The depth of the tree node corresponds directly with scale: the tree root at $s = 1$ corresponds to the entire dataset, $C_{1,1} := X$, at the scale of the diameter of $X$ (i.e., at the coarsest data scale), and each child of a given tree node always corresponds to a subset of the parent's subset (i.e., each $C_{s,p} \subset C_{s-1,p'}$ for some $p'$).  Furthermore, the diameter of each child $C_{s,p}$ is required to be at most a fraction (e.g., one half) of the diameter of its parent $C_{s-1,p'}$ \textit{unless} the parent simply replicates itself at the next scale (i.e., unless $C_{s,p} = C_{s-1,p'}$).  It is this last requirement that enforces the general correspondence between depth in the tree, and spatial scale.  Finally, for any fixed scale $s$, $\{C_{s,1}, \dots, C_{s,P_s}\} $ forms a partition of the dataset $X$.

Then, the following components are computed for each tree node of the GMRA approximation:
\begin{enumerate}

\item A center, $\vec{c}_{s,p} \in \mathbb{R}^D$, that represents the general spatial position of the data in $C_{s,p}$.

\item An orthogonal geometric scaling matrix, $\Phi_{s,p} \in \mathbb{R}^{D \times d}$, whose columns span a $d$-dimensional linear approximation to the re-centered data, \mbox{$(C_{s,p} - \vec{c}_{s,p})$}.  

\item A wavelet constant for $C_{s,p}$ defined by
  \begin{equation*}
    w_{s,p} := \left(I- \Phi_{s-1,p'}\Phi^T_{s-1,p'}
    \right)\cdot(\vec{c}_{s,p}- \vec{c}_{s-1,p'}) \in \mathbb{R}^D,
  \end{equation*}
  where $C_{s-1,p'}$ is the parent subset containing $C_{s,p}$, and an orthogonal wavelet basis matrix, $\Psi_{s,p}$, whose columns span the projection of $\textrm{column\_span}(\Phi_{s,p})$ onto the $(D - d)$-dimensional subspace given by $\left( \textrm{column\underline{ }span}(\Phi_{s-1,p'}) \right)^{\perp} \subset \mathbb{R}^D$. The approximation to $\vec{x} \in X \cap C_{s,p}$ at scale $s$ is
  \begin{equation*} 
    \vec{x}_s := \Phi_{s,p}\Phi^T_{s,p}(
    \vec{x}- \vec{c}_{s,p})+\vec{c}_{s,p}.
  \end{equation*} 
  The wavelet bases and constants allow data points in $X$ to be compactly represented in terms of the telescoping sum $\vec{x} = \vec{x}_1+ \sum^{\infty}_{s = 2} (\vec{x}_{s}- \vec{x}_{s-1})$, representing the differences between the approximations of $\vec{x}$ on consecutive scales as rapidly decaying updates \cite{ALLARD2012435}.
\end{enumerate}

This construction satisfies the assumptions in \cite{maggioni2016multiscale}, giving probabilistic bounds on the performance of the GMRA approximation under mild assumptions on the underlying distribution of the data.

\section{Updating Manifolds}
A benefit of the non-linear manifold constructed described in \cref{sec:gmra} is the simplicity of the update procedure for an existing non-linear manifold given more data. In particular, the covertree structure allows for the easy insertion of data \cite{covertree}, and more importantly, preserves the existing partitioning of the dataset at each scale $s$
unless the manifold changes drastically (requiring additional partitions), in which case the low-dimensional non-linear manifold is not sufficiently accurate, or an outlier data point has been inserted.
Specifically, if the covertree partitioning of a data set gives (at any fixed scale $s$) $\{C_{s,1}, \dots, C_{s,P_s}\} $, then data insertion results in a partitioning of a data set (at scale $s$) $\{\hat{C}_{s,1}, \dots, \hat{C}_{s,P_s}\} $ with $C_{s,i} \subset \hat{C}_{s,i}$ $\forall i \in P_s$ and $\forall s$. Before proceeding, we review a set of sequential update rules for adding data to an SVD decomposition \cite{BRAND200620} 
before applying this approach to update a non-linear manifold constructed using GMRA.

\subsection{Updating an SVD decomposition}
\label{svd_update}
First, recall that computing a rank--$r$ SVD
factorization of $\mathbf{C}\in\mathbb{R}^{p\times q}$ is expensive,
requiring $\mathcal{O}(p\,q\,r)$ operations. Suppose that a rank--$r$
approximation to $\mathbf{C}$, denoted
$\mathbf{C}_r= \mathbf{U\,S\,V}^\top$, is already available.  Here,
$\mathbf{U} \in \mathbb{R}^{p \times r}$,
$\mathbf{S} \in \mathbb{R}^{r\times r}$, and
$\mathbf{V} \in \mathbb{R}^{q\times r}$.  Let
$\mathbf{A} \in \mathbb{R}^{p\times c}$ and
$\mathbf{B} \in \mathbb{R}^{q\times c}$. We seek to modify this thin
SVD factorization when $\mathbf{C}$ is updated,
\begin{align}
  \mathbf{C}_r + \mathbf{A}\,\mathbf{B}^\top,
  \label{eqn:update_eqn}
\end{align}
notated here as an additive update. Using \cite{BRAND200620}, SVD factorization of \cref{eqn:update_eqn} can be computed as follows with
$\mathcal{O}(p\,r + r^3)$ operations.
    \begin{align}
     \mathbf{C}_r +\mathbf{A}\mathbf{B}^\top =
     \left(\begin{bmatrix} \mathbf{U} & \mathbf{P} \end{bmatrix}\mathbf{U}'\right)
     \mathbf{S}'
     \left(\begin{bmatrix} \mathbf{V} & \mathbf{Q} \end{bmatrix}\mathbf{V}'\right)^\top
    \label{eq:brand_final}
    \end{align}  
    \text{where,}\\
    $\mathbf{P}$ -- orthogonal basis for $(\mathbf{I}- \mathbf{U}\,\mathbf{U}^\top)\,\mathbf{A}$

    $\mathbf{Q}$ -- orthogonal basis for $(\mathbf{I}-\mathbf{V}\,\mathbf{V}^\top)\,\mathbf{B}$

    $\mathbf{U}'$  and  $\mathbf{V}'$  diagonalize $\mathbf{K} \in \mathbb{R}^{(r+c)\times (r+c)} $, $\mathbf{U}'^\top\mathbf{K}\mathbf{V}' =\mathbf{S}'$
    \begin{align*}
    \mathbf{K}
    = 
    \begin{bmatrix} \mathbf{S} & \mathbf{0} \\ \mathbf{0} & \mathbf{0} \end{bmatrix}+
    \begin{bmatrix} \mathbf{U}^\top\,\mathbf{A} \\ \mathbf{R}_A \end{bmatrix}
    \begin{bmatrix} \mathbf{V}^\top\,\mathbf{B} \\ \mathbf{R}_B \end{bmatrix}^\top
    \end{align*} 
    
    \hbox{$\mathbf{R}_A = \mathbf{P}^\top\,(\mathbf{I}-\mathbf{U}\,\mathbf{U}^\top)\,\mathbf{A}$,\,
    $\mathbf{R}_B = \mathbf{Q}^\top\,(\mathbf{I}-\mathbf{V}\,\mathbf{V}^\top)\,\mathbf{B}$}

\subsection{Updating Covariance Matrix}
Let $\mathbf{X} \in \mathbb{R}^{D\times n}$ be the original set of
observations and $\hat{\mathbf{X}} = \begin{bmatrix} \mathbf{X} &
\mathbf{c} \end{bmatrix}$ be the data set appended by a new observation, 
$\mathbf{c}\in \mathbb{R}^{D\times 1}$. 

The updated covariance matrix $\cov(\hat{\mathbf{X}})$ can be
written as a rank--$1$ update of $\cov(\mathbf{X})$,
\begin{align}
\cov(\hat{\mathbf{X}}) = \left(\frac{n-1}{n}\right)\cov(\mathbf{X}) +
\frac{1}{(n+1)}(\bar{\mathbf{x}} - \mathbf{c})(\bar{\mathbf{x}} - \mathbf{c})^\top
\label{eq:cov_rank1}
\end{align}

Notice that \Cref{eq:cov_rank1} is in the format of \Cref{eqn:update_eqn}. Therefore, using \Cref{eq:brand_final} we can perform the incremental update of SVD without recomputing a new covariance matrix. Similarly, a rank--$m$ update can be performed as follows:

Let $\mathbf{X} \in \mathbb{R}^{D\times n}$ be the data set of
interest and $\hat{\mathbf{X}} = \begin{bmatrix} \mathbf{X}
& \mathbf{C} \end{bmatrix}$ with $\mathbf{C}\in \mathbb{R}^{D\times
m}$. Here, we have added $m$ new observations. The covariance matrix
of $\hat{\mathbf{X}}$ can be written in terms of the covariance matrix
of $\mathbf{X}$,

\begin{align}
    \cov(\hat{\mathbf{X}}) = a\,\cov(\mathbf{X}) + b\,\mathbf{B}\,\mathbf{B}^\top
\end{align}
where 
\begin{align*}
    a &= \left(\frac{n-1}{n+m-1}\right), \,
    b = \frac{1}{(n+m-1)}, \\
    \mathbf{B} &= \left[(\mathbf{x}_{n+1}-\bar{\mathbf{c}}),\ldots,(\mathbf{x}_{n+m}-\bar{\mathbf{c}}),\,\sqrt{\frac{n\,m}{(n+m)}}
  (\bar{\mathbf{x}} - \bar{\mathbf{c}})\right].
\end{align*}

The proof of rank--$m$ update is given in \ref{rank_m_update}, and rank--$1$ update is a special case of rank--$m$ update where $m = 1$.

\subsection{Incremental GMRA Algorithm}
\label{subsec:inc_gmra_alg}
We are now ready to summarize the incremental GMRA algorithm, \cref{alg:incremental_gmra}, before making a few remarks.

\begin{algorithm}[htbp]
\caption{Incremental GMRA Algorithm}
\label{alg:incremental_gmra}
\begin{algorithmic}[1]
\Statex \textbf{Input:}
\Statex \qquad $\mathcal{T}$ -- covertree structure
\Statex \qquad $\mathcal{G}$ -- GMRA structure
\Statex \qquad $\mathbf{c}$ -- a new data point from the data stream 
\Statex \textbf{Output:} 
\Statex \qquad $\mathcal{T}_\text{new}$ -- updated covertree structure
\Statex \qquad $\mathcal{G}_\text{new}$ -- incrementally updated GMRA structure
\Statex \textbf{Steps:}
\State $s = \mathcal{T}.$maxlevel
\State  $[\; \mathcal{T}_\text{new}, Z\;] = \textsc{insert}(\mathbf{c},\mathcal{T},s)$ 

\For {cluster in $Z$}
    \State Update PCA basis $\phi_{s,p}$ using \cref{eq:brand_final},

\EndFor
\end{algorithmic}
\end{algorithm}

\begin{remark}
  In line 2 of \cref{alg:incremental_gmra}, the \textsc{insert} function is the Cover Tree insert function that have been slightly modified to return the set of clusters, $Z$,
  consists of the leaf cluster containing the new data point to be
  added, as well as all the parents. 
\end{remark}

\begin{remark}
  The for loop lines 3--5 in \cref{alg:incremental_gmra}, is the
  ``price'' we pay for benefits afforded by the multiscale GMRA
  structure. In addition to updating the basis in the leaf cluster, we
  also need to update all the basis in all the parents.
\end{remark}

\begin{remark}
  If we only care about generating the non-linear manifold that
  supports the data, then \cref{alg:incremental_gmra} can be
  simplified in line 2 to only identify the leaf cluster containing
  $\mathbf{c}$, and if they exist, any other new clusters created as a
  result of $\mathbf{c}$.
\end{remark}

\begin{remark}
    A key parameter in our algorithm is a threshold value used to determine whether to split a cluster.  Specifically, for a cluster at level $s$ in the cluster map, we measure
\begin{align}
    \label{eqn:mse_jk}
    \text{MSE}_{s,p} = \frac{1}{|C_{s,p}|}\sum_{x_i \in C_{s,p}}
    \left\| \mathbf{x}_i- 
    \hat{\mathcal{P}}_{\mathcal{M}_s} 
    (\mathbf{x}_i) 
    \right\|^2,  
\end{align}
where $ \hat{\mathcal{P}}_{\mathcal{M}_s} $ is the projection onto the $s$-level construction of the non-linear manifold. In our numerical experiments, we set the threshold value for \cref{eqn:mse_jk} to be $\epsilon = 0.1$ unless otherwise noted.

A second parameter in our cluster map construction and incremental algorithm is $M$, the minimum number of samples needed in a cluster before a split is considered. We set $M = 30$ as the threshold value of the minimum number of samples in a cluster.
\end{remark}

\section{Principal Angles between Subspaces}
\label{sec:angles}
To detect potential drifts in the support of the low-dimensional manifold, we need tools to understand how the underlying low-dimensional manifold is changing. A surprisingly overlooked tool is the measurement of angles between subspaces \cite{afriat_1957,MR1165446}. 

\subsection{Angle between two vectors}
To introduce these ideas, we recall a simple idea from geometry, that the cosine of the angle between two unit vectors (anchored at the origin) satisfies
\begin{align*}
  \cos{\theta} = \vec{x}^\top \vec{y},
\end{align*}   
where $\|\vec{x}\|_2 = \|\vec{y}\|_2 = 1$.

\subsection{Angles between two subspaces}
The higher-dimensional version extends as follows.
Let $U_1$ and $U_2$ be subspaces of $\mathbb{R}^D$, with 
$$\dim{U_1} = d_1 \le \dim{U_2} = d_2.$$
Then, the principal angles between $U_1$ and $U_2$,
$$ 0 \le \theta_1 \le \theta_2 \le \cdots \le \theta_{d_1} \le \frac{\pi}{2},$$
satisfy (recursively) for $i = 1, 2, \ldots, d_1$,
\begin{align}
\nonumber
\cos{\theta_i} 
&= \max \left\{ \frac{\langle \; \vec{x} \;,\; \vec{y}\; \rangle}{\|\vec{x}\|\,\|\vec{y}\|} : 
\begin{matrix}
\vec{x} \in \spn{(U_1)}, & \vec{x}^\top \vec{x}_k = 0,  \\
\vec{y} \in \spn{(U_2)}, &  \vec{y}^\top \vec{y}_k = 0, \\
&  \forall k < i
\end{matrix} 
\right\} \\
&=  
\frac{\langle \; \vec{x}_i \;,\; \vec{y}_i\; \rangle}{\|\vec{x}_i\|\,\|\vec{y}_i\|}. 
\label{eqn:principal_angle}
\end{align}
By construction, one can show that
\begin{align*}
\cos{\theta_i} = \sigma_i, 
\quad
i = 1,2,\ldots d_2,
\end{align*}
where $\sigma_i$ are the singular values of the matrix $U_2^\top U_1$ \cite{MR348991}.

\section{Error Analysis}
In this section we study the errors due to incremental updates using Brand's method. Brand's procedure, described in \cref{svd_update}, provides an efficient way to update the rank--$d$ SVD factorization of the covariance (or data)
matrix, given new observations. However, the resulting modified
factorization may not be the same as recomputing the covariance matrix
of the combined data set, and then forming its rank--$d$
approximation as shown in \Cref{eqn:Rd1,eqn:Rd2}. We discuss the potential differences in singular values ${\mathbf{S}}_d$ and $\tilde{\mathbf{S}}_d$ in \Cref{lemma:singular_value_update}. The proof is given in \ref{appendix:lemma1_proof}.
\\
\begin{lemma}
\label{lemma:singular_value_update}
Let $\mathbf{C}=\mathbf{U\,\Sigma\,U^\top}$ be the covariance matrix
associated with the original data, and
$\mathbf{C}_d=\mathbf{U}_d\,\mathbf{\Sigma}_d\,\mathbf{U}_d^\top$ be the
corresponding rank--$d$ approximation. Consider the SVD of the additive updates 
\begin{align}
\left(a\,\mathbf{C} + b\,\mathbf{B}\,\mathbf{B}^\top\right)_d
= \mathbf{R}_d\,\mathbf{S}_d\,\mathbf{R}_d^\top ,
\label{eqn:Rd1}
\end{align}
and
\begin{align}
\left(a\,\mathbf{C}_d + b\,\mathbf{B}\,\mathbf{B}^\top\right)_d
= \tilde{\mathbf{R}}_d\,\tilde{\mathbf{S}}_d\,\tilde{\mathbf{R}}_d^\top.
\label{eqn:Rd2}
\end{align}

Then
\begin{align}
\nonumber
\left|\sigma_i(a\,\mathbf{C} + b\,\mathbf{B}\,\mathbf{B}^\top) - \sigma_i(a\,\mathbf{C}_d + b\,\mathbf{B}\,\mathbf{B}^\top)\right| \\
\le 
 a\,\sigma_{d+1}(\mathbf{C}) + b\,\sigma_1^2(\mathbf{B}). 
\end{align}
\end{lemma}

Numerical experiments we performed to explore the difference between singular values for a generic (symmetric) matrix revealed that the upper bound in \Cref{lemma:singular_value_update} is not a tight bound. However, the numerical experiments yielded another useful variation for the upper bound given in \cref{lemma:singular_value_update} which is tighter. This is given in \cref{prop1}.  
\\
\begin{proposition}
\label{prop1}
Let $\mathbf{C}=\mathbf{U\,\Sigma\,U^\top}$ be the covariance matrix
associated with the original data $\mathbf{X}$, and
$\mathbf{C}_d=\mathbf{U}_d\,\mathbf{\Sigma}_d\,\mathbf{U}_d^\top$ be the
corresponding rank--$d$ approximation. Consider the SVD of the additive updates in \Cref{eqn:Rd1,eqn:Rd2}. Then,

\begin{align}
  \|\mathbf{S}_d - \tilde{\mathbf{S}}_d\|_2 \sim \mathcal{O}
  \left(
   \sigma_{d+1}(\mathbf{C})\, \|\mathbf{B}\|_F^4\right)
\end{align}
\end{proposition}

More interestingly, we wish to compare how the PCA basis specified by
$\mathbf{R}_d$ and $\tilde{\mathbf{R}}_d$ in \cref{eqn:Rd1,eqn:Rd2}
differ. Using Theorem 3 \cite{zwald2005convergence}, we approach the lemma \ref{lem:perturb_angle_bound}.  
\\
\begin{lemma}
\label{lem:perturb_angle_bound}
Let $a\,\sigma_{d+1}(\mathbf{C}) < \left(\sigma_d(\mathbf{C})+\sigma_1^2(\mathbf{B})\right)/4$. The principal angle(s) between the subspaces specified by
$\mathbf{R}_d$ and $\tilde{\mathbf{R}}_d$ in \cref{eqn:Rd1,eqn:Rd2}
satisfies
\begin{align}
\nonumber
\|\tilde{\mathbf{R}}_d\,\tilde{\mathbf{R}}_d^\top - 
    {\mathbf{R}}_d\,{\mathbf{R}}_d^\top\|_2 = \|\sin{\mathbf{\Theta}}\|_2 \\
    \leq 
    \frac{2\,(n-1)\,\sigma_{d+1}(\mathbf{C})}{(n-1)\,\sigma_d(\mathbf{C})+\sigma_1^2(\mathbf{B})}
\end{align}
where
\begin{align*}
   \mathbf{\Theta} = \text{diag}(\theta_1,\ldots, \theta_d) 
\end{align*}
\end{lemma}
The subspaces represented by $\tilde{\mathbf{R}}_d$ and ${\mathbf{R}}_d$ yields relatively small angular separations when the $\sigma_{d+1}(\mathbf{C})$ is relatively small. Small $\sigma_{d+1}(\mathbf{C})$ values indicate that the discarded information is not significant and, the original structure is well-captured by the rank--$d$ approximation.

\section{Numerical Experiments}
\label{sec:experiments}
The proposed incremental GMRA algorithm will be tested under the scenario where training data is sampled from a smooth manifold $\mathcal{M}$. However, the training samples do not capture {\em all} the features of the manifold. We wish to verify that streaming in more data and incrementally updating the GMRA structure provides an updated manifold with all the features.

The hypothesis for the experiments is that the manifold approximation improves as we progressively stream in more samples. To quantify this, we will measure the Mean Squared Error (MSE) of the training and streamed data, measured as
\begin{align}
\text{MSE} = \frac{1}{(n+m)} \sum_{i=1}^{n+m} \left\| \mathbf{x}_i - \hat{\mathcal{P}}_\mathcal{M}(\mathbf{x}_i)  \right\|_2^2,
\end{align}
where $\mathbf{x}_i$ is observation $i$, $n$ is the number of training observations, $m$ is the number of streamed observations, and $\hat{\mathcal{P}}_\mathcal{M}$ is the projection onto the (incremented) GMRA generated non-linear manifold.

\subsection{Swiss Roll Manifold}
\label{subsec:case1_swiss_roll}
First, we randomly sampled data from a Swiss roll: a manifold embedded in $\mathbb{R}^3$ but with an intrinsic dimension in $\mathbb{R}^2$. Swiss roll is widely used in the literature to benchmark manifold learning algorithms. Swiss roll data can be generated by using the following definition.
\begin{align*}
    x(t) = t\,\cos{(t)},\quad y = h,\quad z(t) = t\,\sin{(t)}
\end{align*}
For this experiment we use the Python scikit--learn package \cite{scikit-learn} to generate the data; which uses $t$ from Uniform($1.5\pi,4.5\pi$) and $h$ from Uniform($0, 21$) \cite{marsland2014machine}.

We randomly sample 50,000 data points from the Swiss roll manifold and use 500 of them as the training set to create the initial GMRA structure. Then we stream in the remaining 49,500 sample points, incrementally updating the GMRA structure. 

\begin{figure}[htbp]
  \begin{center}
    \subfigure[Initial Nonlinear Manifold Approximation (using training data)]{
      \includegraphics[width=0.32\textheight,height=0.30\textheight]{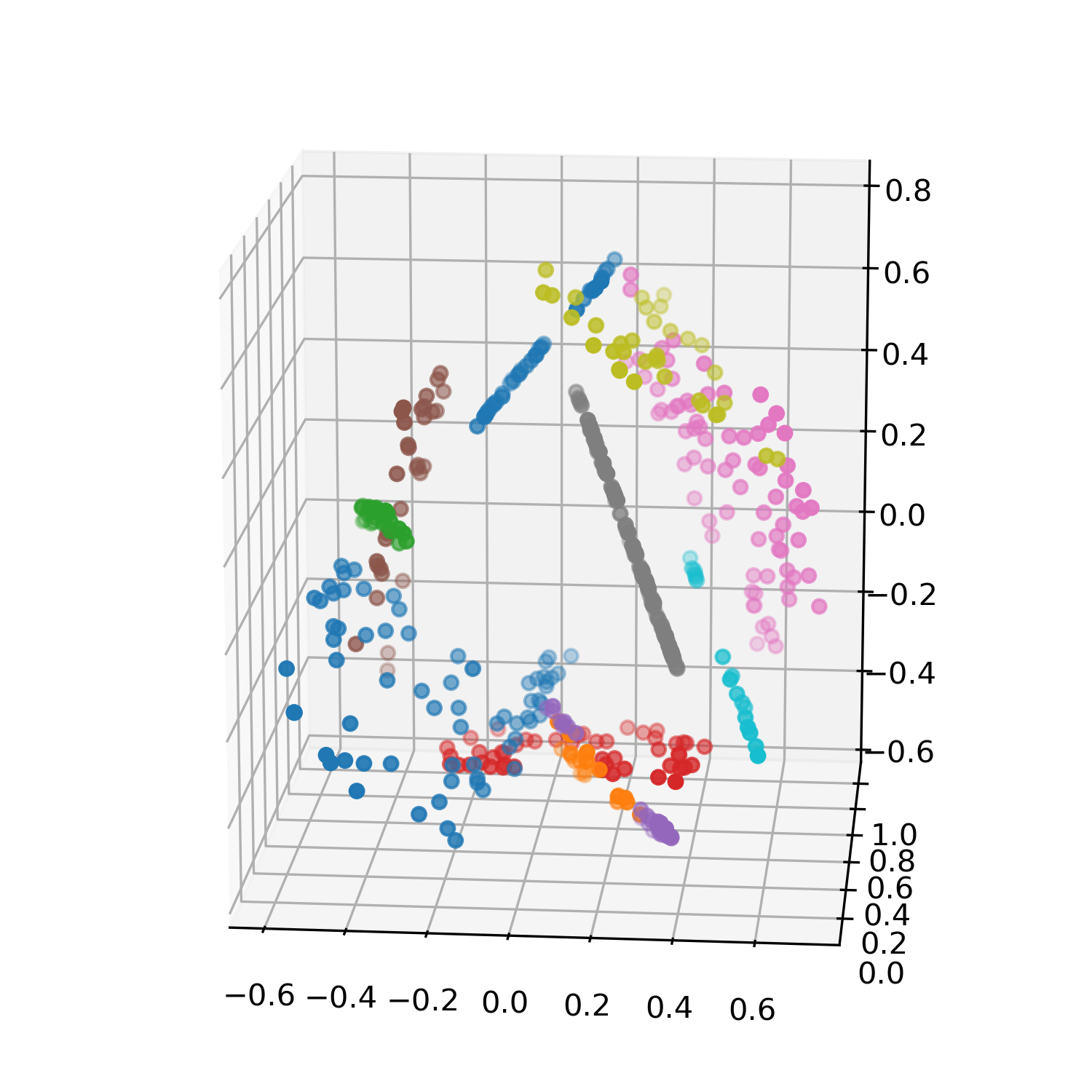}
      \label{fig:swissroll_train_fit}
    } 
    \subfigure[Incremented Manifold Approximation ($2000$ data points added)]{
    \includegraphics[width=0.32\textheight,height=0.30\textheight]{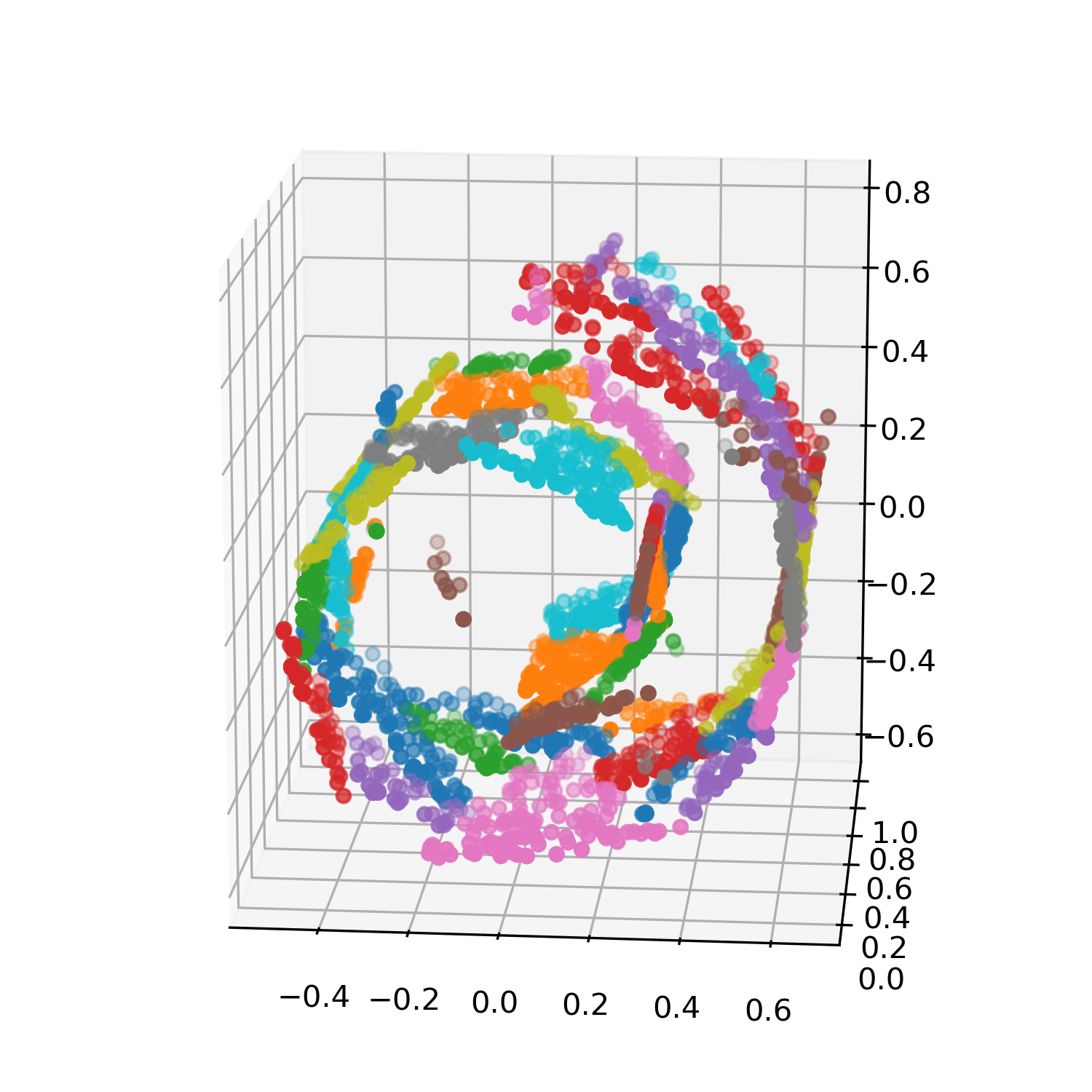}
    \label{fig:swissroll_stream2000}
    }  \\
    \subfigure[Incremented Manifold Approximation (remaining $47,500$ data points added)]{
    \includegraphics[width=0.32\textheight,height=0.30\textheight]{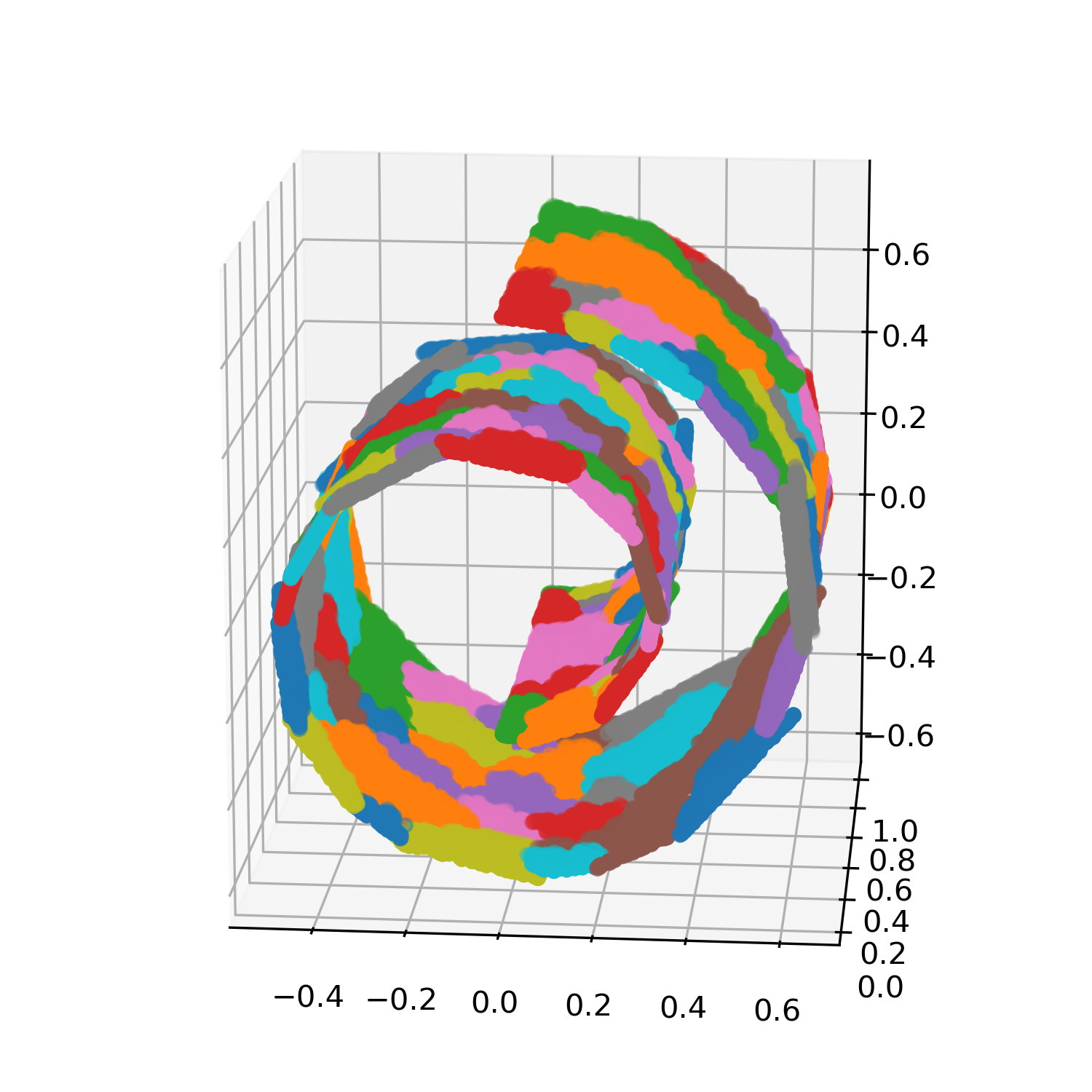}
    \label{fig:swissroll_stream10k}
    }  
    \subfigure[Ground Truth]{
    \includegraphics[width=0.32\textheight,height=0.30\textheight]{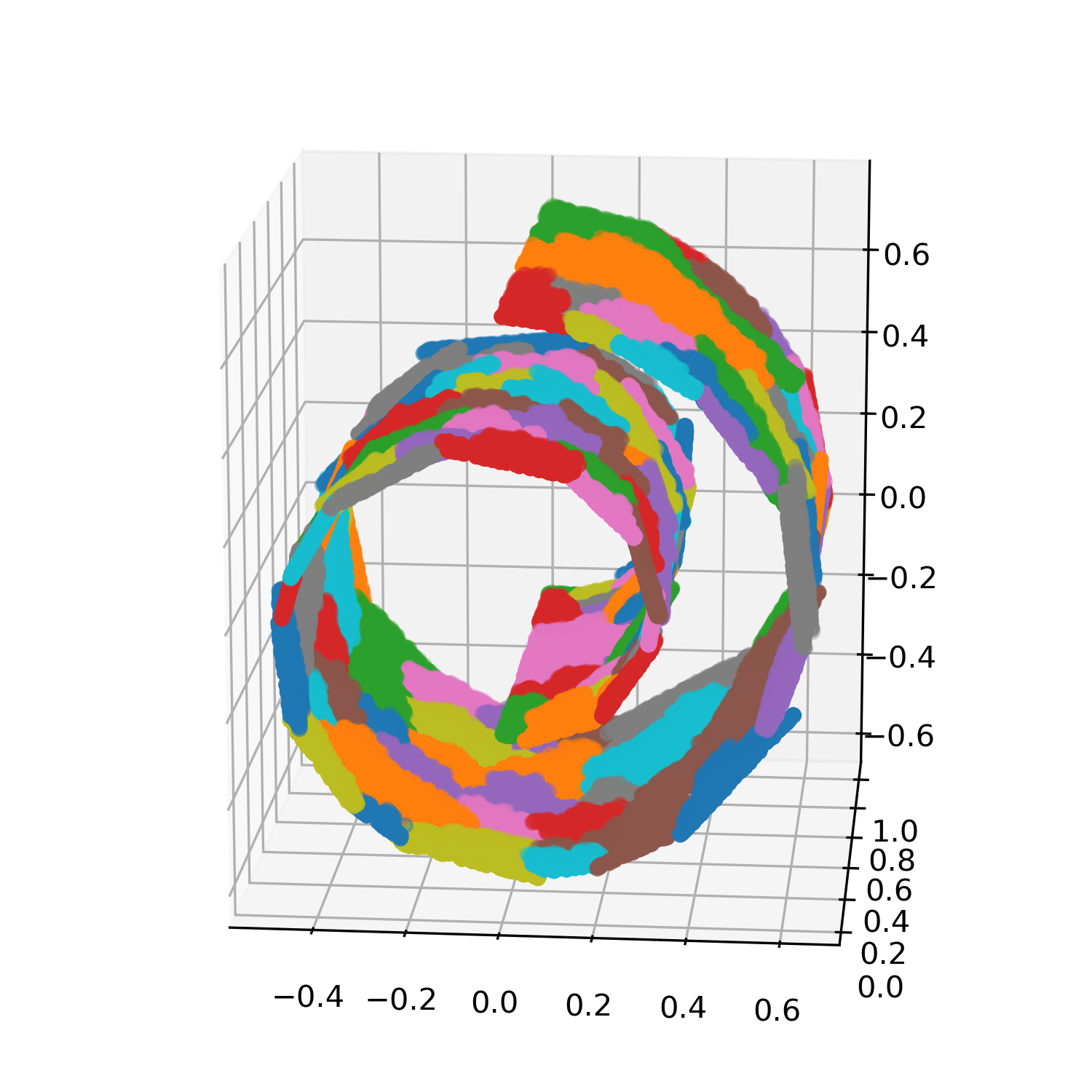}
    \label{fig:swiss_groudtruth}
    }
  \end{center}
  \caption{Figure (a) shows the low-dimensional GMRA approximation for the 500 training data. Figure (b) displays the incrementally updated low-dimensional approximation following the inclusion of the 2000 more samples, and figure (c) shows the final GMRA approximation after streaming in all the remaining samples. Figure (d) -- Low dimensional approximation when all 50,000 sample points are used to create GMRA structure (Ground truth)}
  \label{fig:swiss_roll}
\end{figure}

\begin{figure}[htbp]
    \centering
        \subfigure[MSE vs Sample Size]{
            \includegraphics[width=0.32\textheight,height=0.30\textheight]{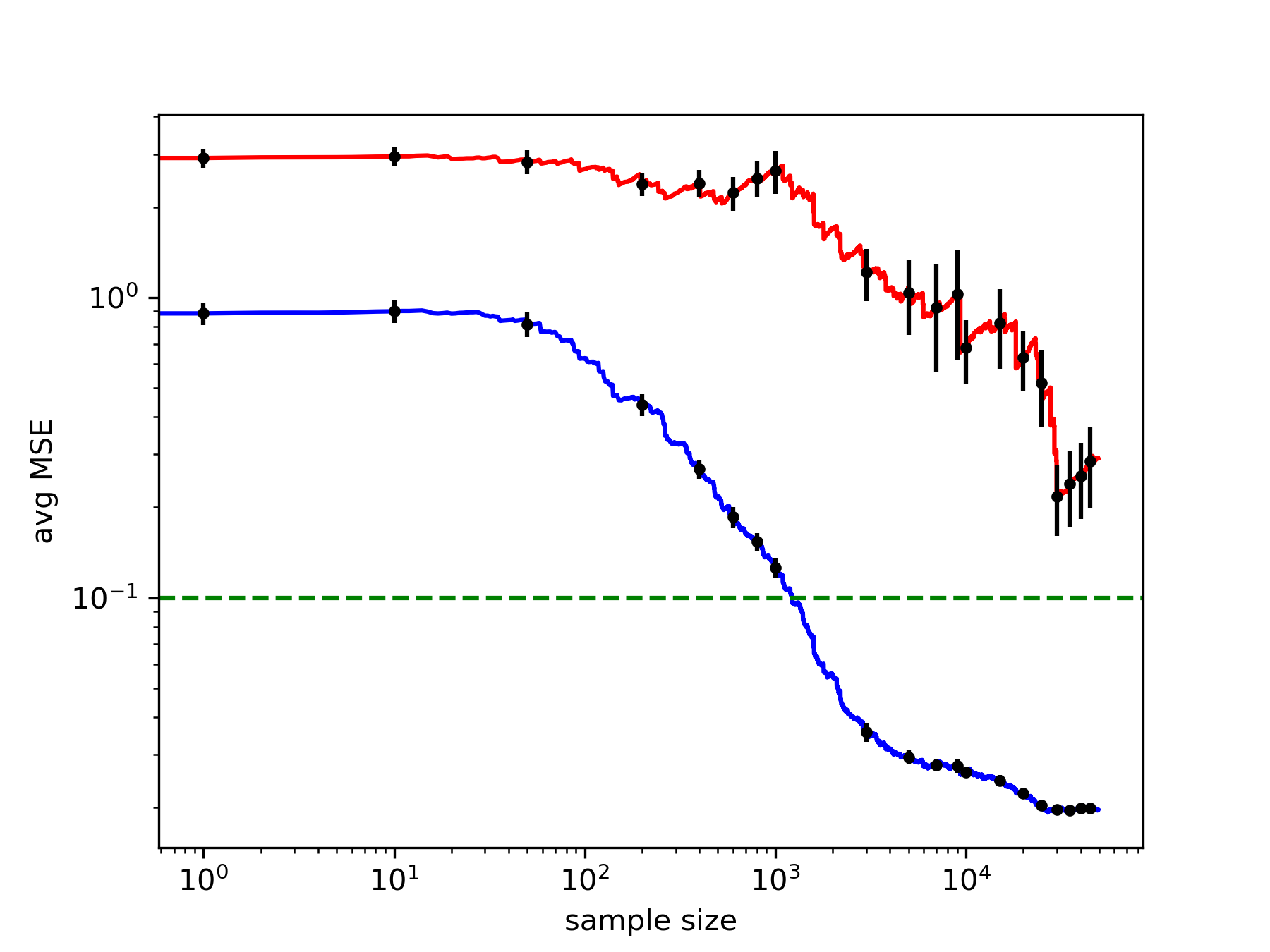}
            \label{fig:swiss_mse_rep}        
        }
        \subfigure[Number of Leaf Clusters vs Sample Size]{
            \includegraphics[width=0.32\textheight,height=0.30\textheight]{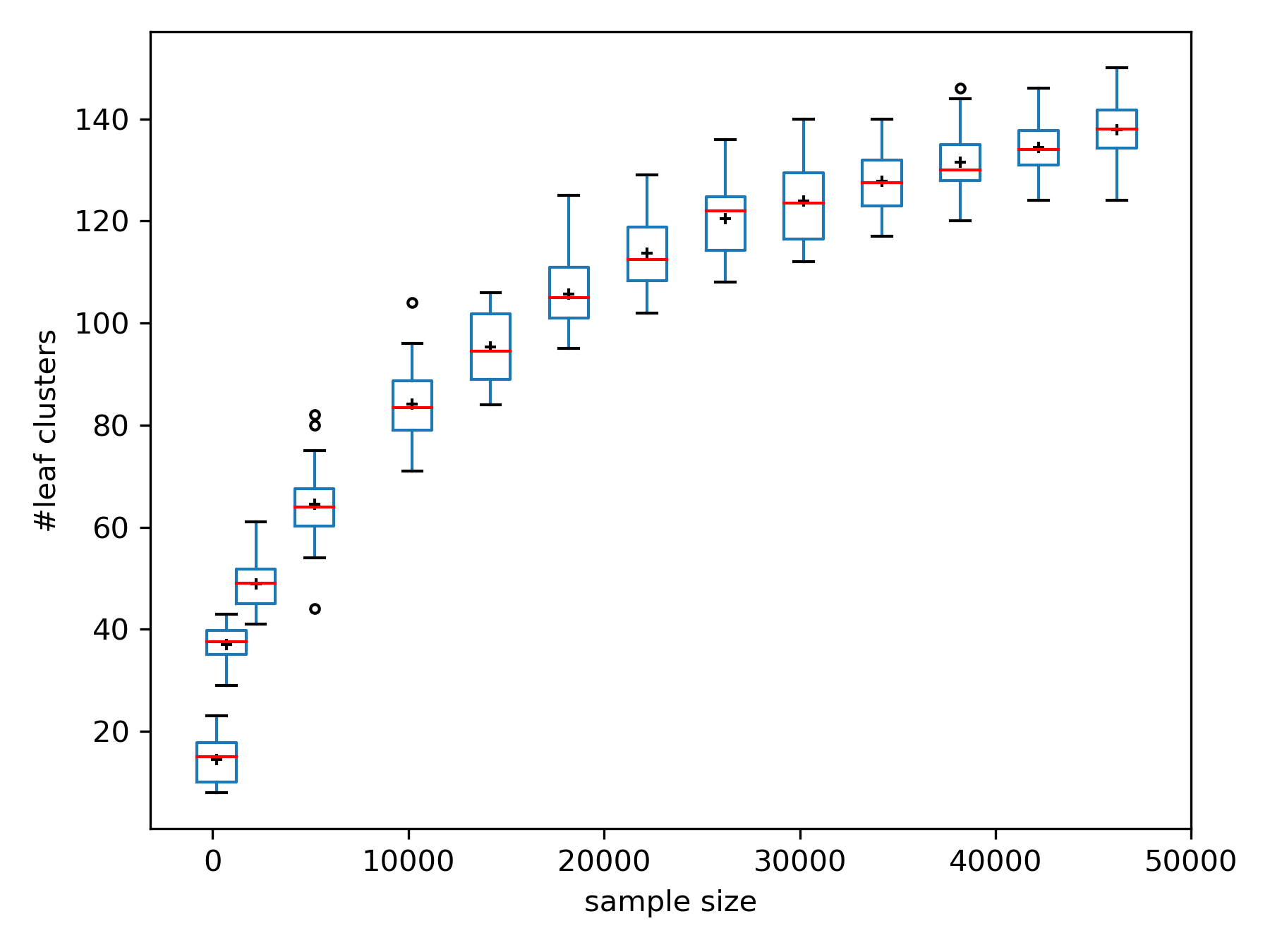}
            \label{fig:swiss_num_clusters_rep}        
        } \\
        \subfigure[Depth vs Sample Size]{
            \includegraphics[width=0.32\textheight,height=0.30\textheight]{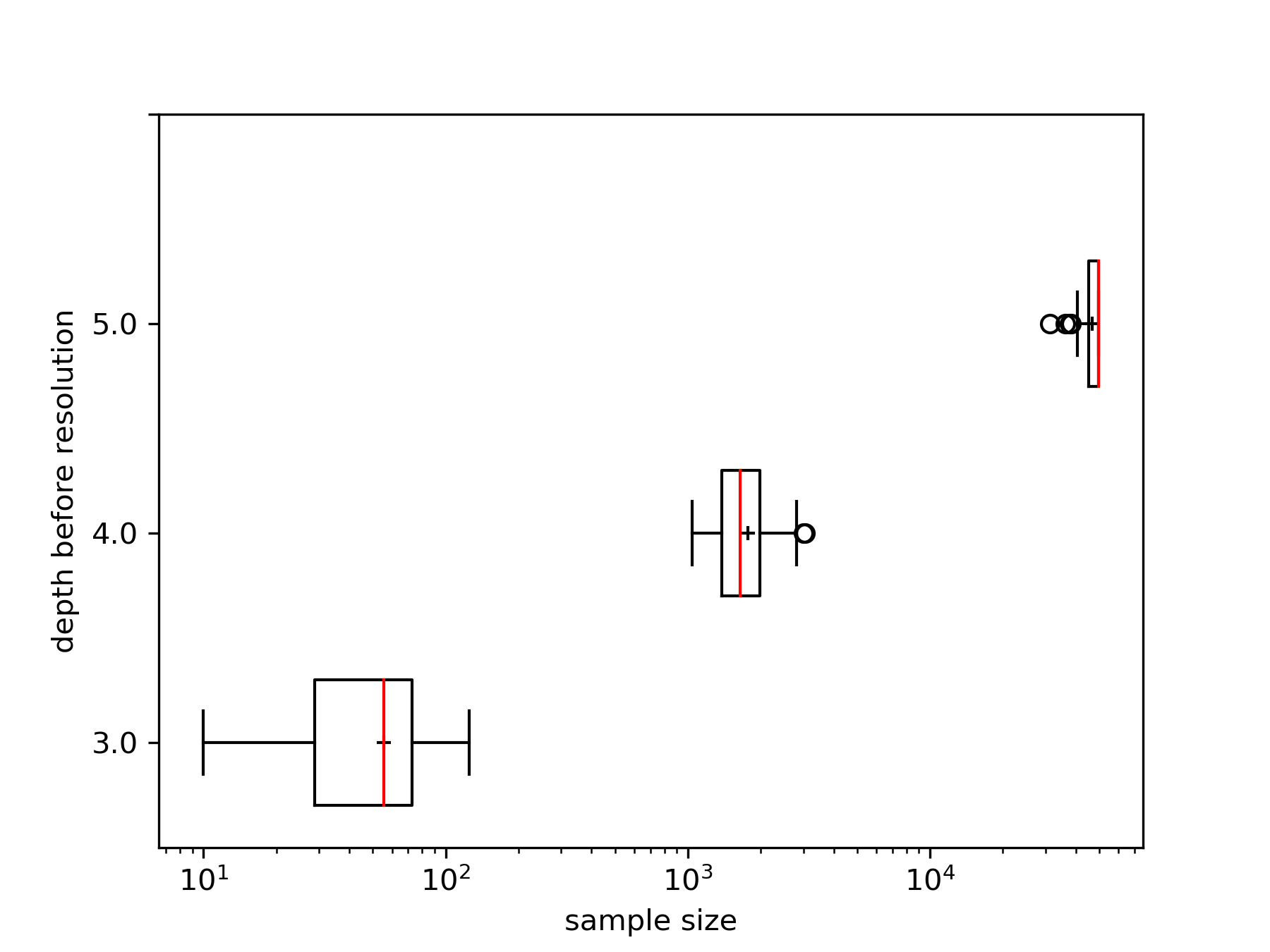}
            \label{fig:swiss_depth_rep}        
        }
        \subfigure[Cluster Size]{
            \includegraphics[width=0.32\textheight,height=0.30\textheight]{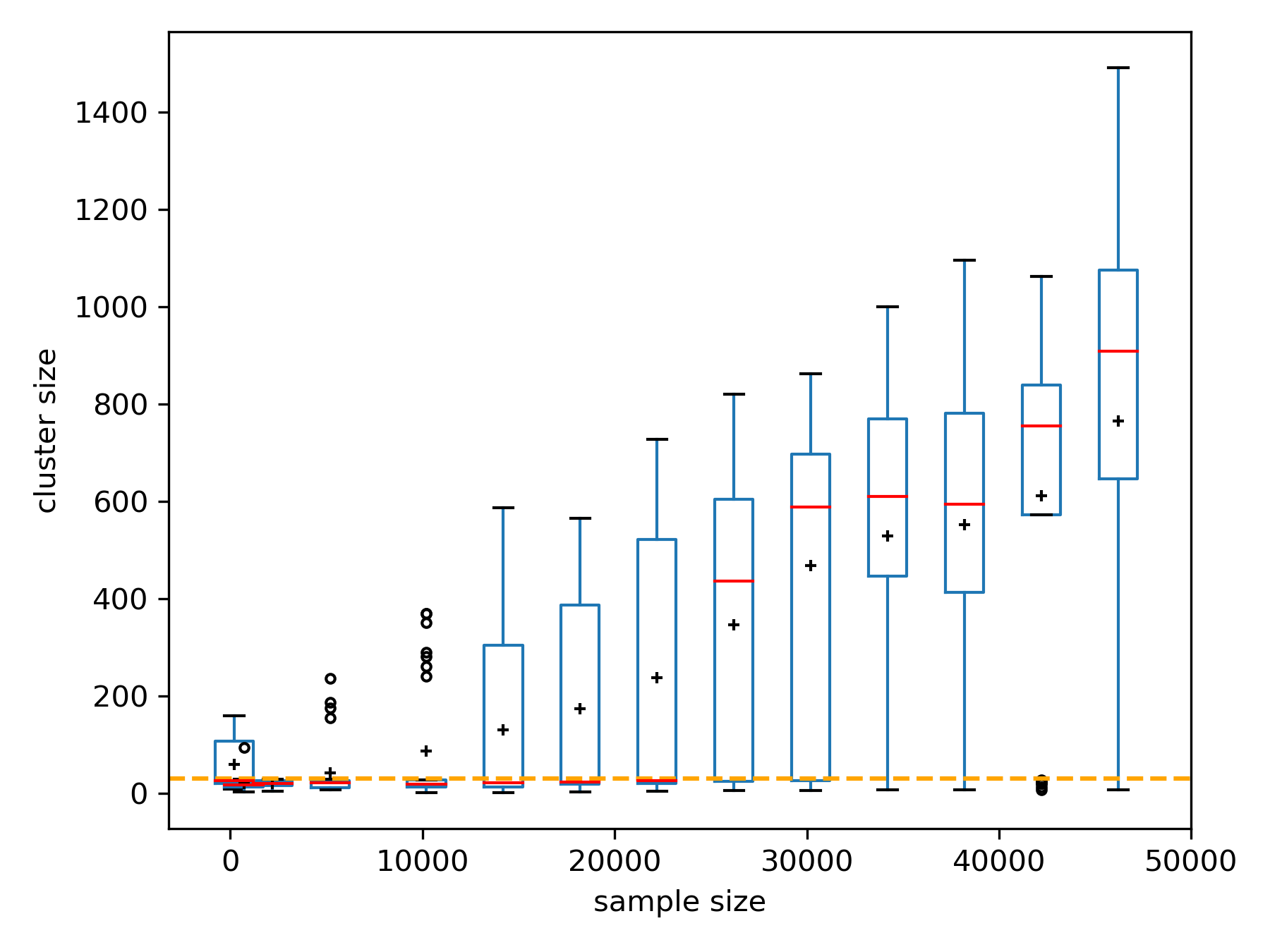}
            \label{fig:swiss_numpts_rep}        
        }                 
    \caption{Figure (a) shows the average mean squared error of the approximation (blue) by sample size for the Swiss Roll data, along with the average maximum MSE (red) at leaf clusters after 30 repeats. The standard error bars are also depicted on each curve. A green horizontal line depicts the threshold level $\epsilon = 0.1$. Figure (b) shows the boxplots of the number of leaf clusters used for the approximation by sample size after 30 repeats. The line bisecting the box represents the median and the $+$ sign within the box represents the mean number of leaf clusters. Figure (c) illustrates the maximum height of the cluster map. Each boxplot in Figure (c) illustrates, on average, how many new samples it takes before the cluster map needs to resolve an additional depth (based on 30 repeats of the experiment). Figure (d) depicts the boxplots of the number of samples within the leaf cluster where the maximum MSE is observed after each increment (the orange horizontal line shows $M = 30$).}
    \label{fig:swiss_mse_num_clusters_rep}
\end{figure}

\Cref{fig:swiss_roll} illustrates the results of this experiment for a single run. \Cref{fig:swissroll_train_fit} shows the GMRA approximation on $500$ training samples and in \cref{fig:swissroll_stream2000,fig:swissroll_stream10k}, we show the incremented GMRA approximations after $2000$ and $50,000$ data points are included in the GMRA approximation. The ground truth is shown in \cref{fig:swiss_groudtruth}. Then the experiment is repeated 30 times and the average of MSE across 30 repeats along with the average of maximum MSE within leaf clusters are shown in \cref{fig:swiss_mse_rep}, the boxplots of number of leaf clusters used to generate the GMRA approximations at each increment are shown in \cref{fig:swiss_num_clusters_rep} along with the depth before a cluster map resolution are shown in \cref{fig:swiss_depth_rep} and the boxplot of number of samples within the leaf cluster where maximum MSE is observed is given in \cref{fig:swiss_numpts_rep}.

The RMSE error metric value for the difference between the ground truth and stream approximation is $5.61 \times 10^{-6}$. 
In \Cref{fig:swiss_mse_rep} we observe that both the average MSE and average maximum MSE stay relatively constant until about 100 increments. Then, the MSE values start to decrease. The average MSE (blue) goes below the threshold of 0.1 after about 1000 increments. However, the maximum MSE observed (red curve) within a leaf cluster is still above the threshold value.

In \cref{fig:swiss_num_clusters_rep} we can see that the rate of increase of the number of leaf clusters is rapid until about $10,000$ increments where it reaches about 80 leaf clusters on average. After $10,000$ increments, the rate of increase of clusters is relatively slower, where it reaches about 140 leaf clusters in total at the end of $50,000$ increments. In \cref{fig:swiss_depth_rep} we observe that the GMRA structure starts with a depth of three and after about 60 increments the depth increases to a value of four. Then, it says until about $2000$ increments before increasing the resolution to a depth of five. In \cref{fig:swiss_numpts_rep}, we can observe that the median number of samples within the leaf cluster where the maximum MSE was observed is less than $M = 30$ for the first $25,000$ increments.

Due to the complexity of the manifold, the cluster map requires several clusters to capture all the features of the Swiss roll manifold. Therefore, the cluster map continues to add several new clusters, as evident from \cref{fig:swiss_num_clusters_rep}. As these new clusters have only a few observations (less than 30), the manifold approximation gets computed using the basis vectors of the parent clusters. This explains the observation we made in \cref{fig:swiss_mse_rep} where we saw that the maximum value of the $\text{MSE}_{s,p}$ (red curve) at the leaf clusters exceeding the threshold of 0.1.

\subsection{Complex Manifold}

Inspired by the work done in \cite{mahapatra2017s} we test the performance of our algorithm in a scenario where 
the true manifold consists of the intersection of two distinct smooth manifolds, but training data is only given on one of the manifolds. Specifically, we consider a manifold that consists of a Swiss roll (similar to the one described in \Cref{subsec:case1_swiss_roll}) intersected with a hyperplane. 

First, we sample $1000$ data points for the training data from the Swiss roll manifold and construct the initial GMRA structure. Then we sample an additional $49,000$ data points from the Swiss roll and $10,000$ samples from the hyperplane and stream them into the initial GMRA structure randomly. We wish to test if the algorithm is capable of identifying and approximating new manifolds that emerge.

\begin{figure}[htbp]
  \begin{center}
      \subfigure[Swiss Roll and Hyperplane Samples]{
      \includegraphics[trim=1.5cm 1.5cm 1.5cm 1.5cm, clip,width=0.32\textheight,height=0.30\textheight]{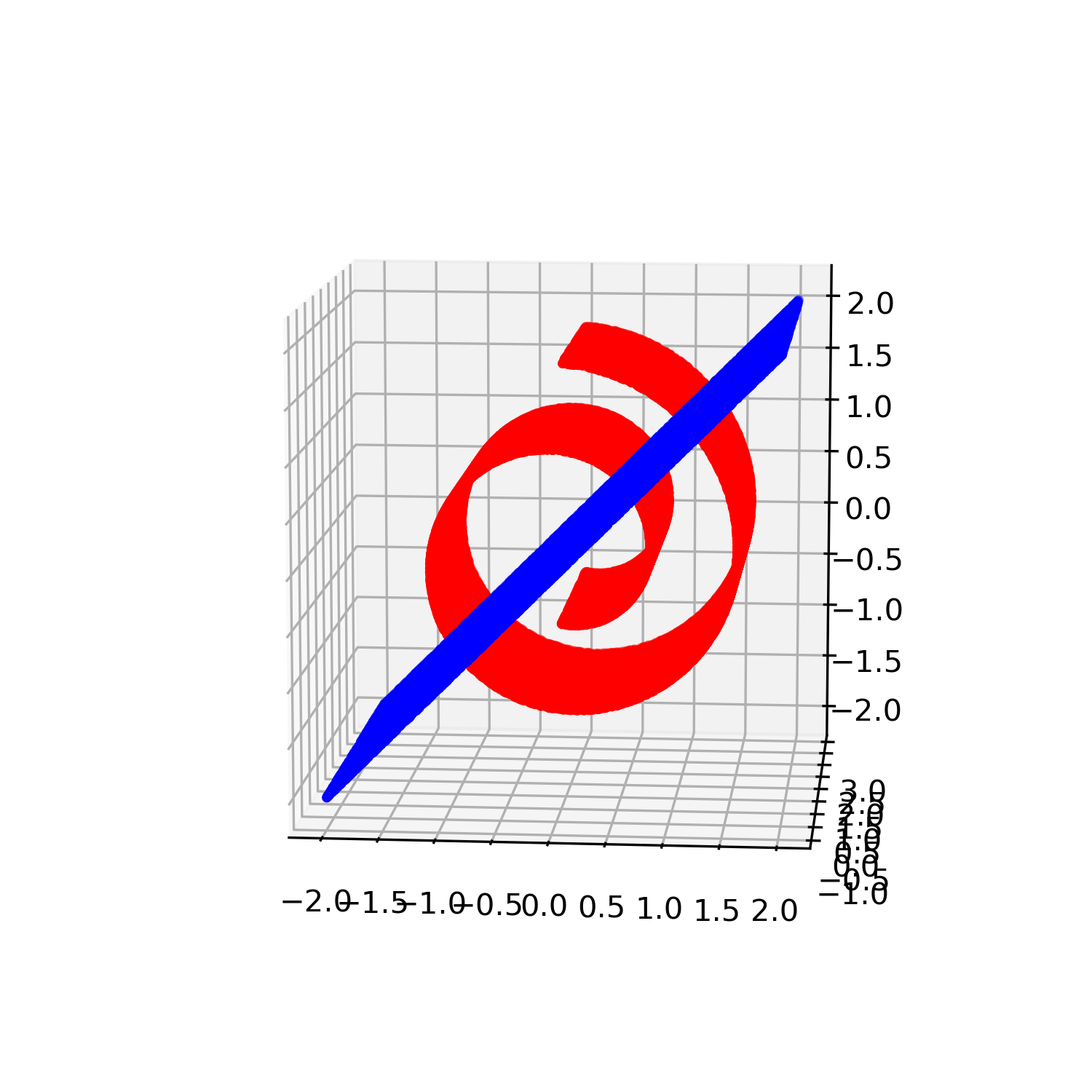}
      \label{fig:swissroll_with_plane}
    }
    \subfigure[Initial Nonlinear Manifold Approximation (using training data)]{
      \includegraphics[width=0.32\textheight,height=0.30\textheight]{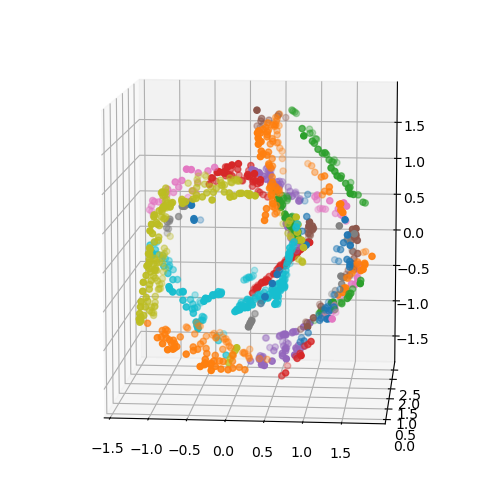}
      \label{fig:srp_train_fit}
    } \\
    \subfigure[Final Incremental GMRA Approximation]{
    \includegraphics[width=0.32\textheight,height=0.30\textheight]{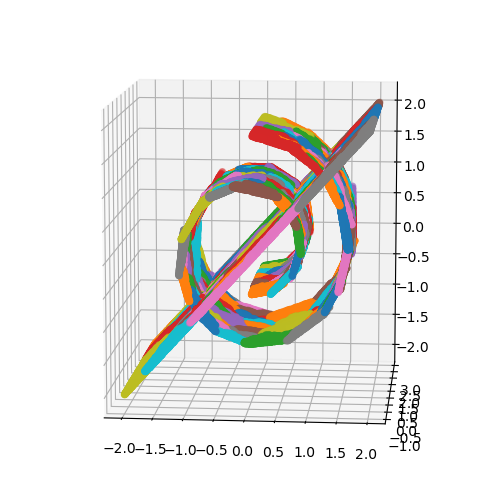}
    \label{fig:srp_stream}
    }
    \subfigure[Ground Truth]{
    \includegraphics[width=0.32\textheight,height=0.30\textheight]{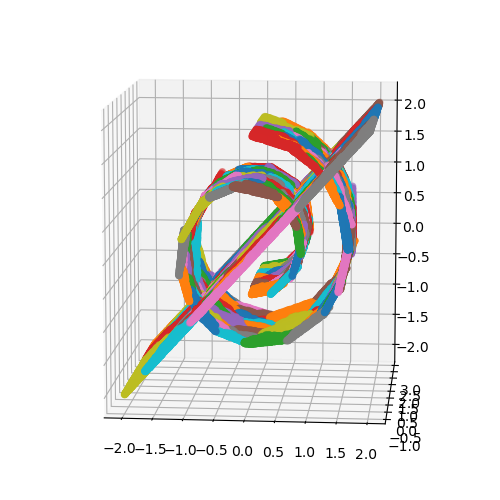}
    \label{fig:srp_grouthtruth}
    }    
  \end{center}
  \caption{Figure (a) shows the Swiss roll with a linear hyperplane intersected. 
  Figure (b) shows the approximation for the 1000 training samples from the Swiss roll manifold. Figure (c) shows the incrementally updated low-dimensional approximation following the inclusion of all 60,000 samples (50,000 from Swiss roll and 10,000 from plane). Figure (d) --GMRA construction for the entire data set of linear hyperplane intersecting a Swiss roll (Ground truth)}
    \label{fig:srp_results}
\end{figure}

\begin{figure}[htbp]
    \centering
        \subfigure[MSE vs Sample Size]{
            \includegraphics[width=0.32\textheight,height=0.30\textheight]{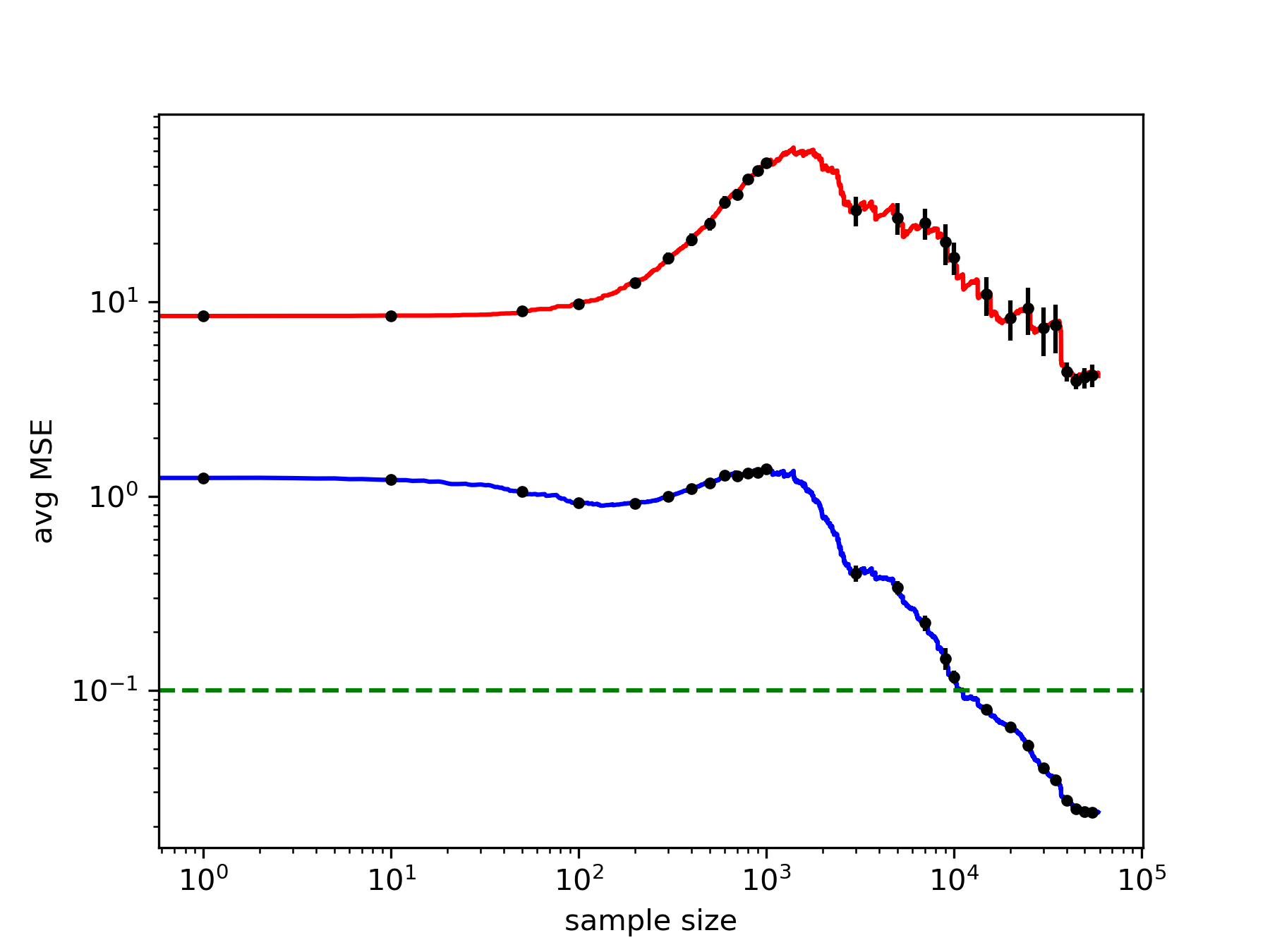}
            \label{fig:srp_mse_rep}        
        }
        \subfigure[Number of Leaf Clusters vs Sample Size]{
            \includegraphics[width=0.32\textheight,height=0.275\textheight]{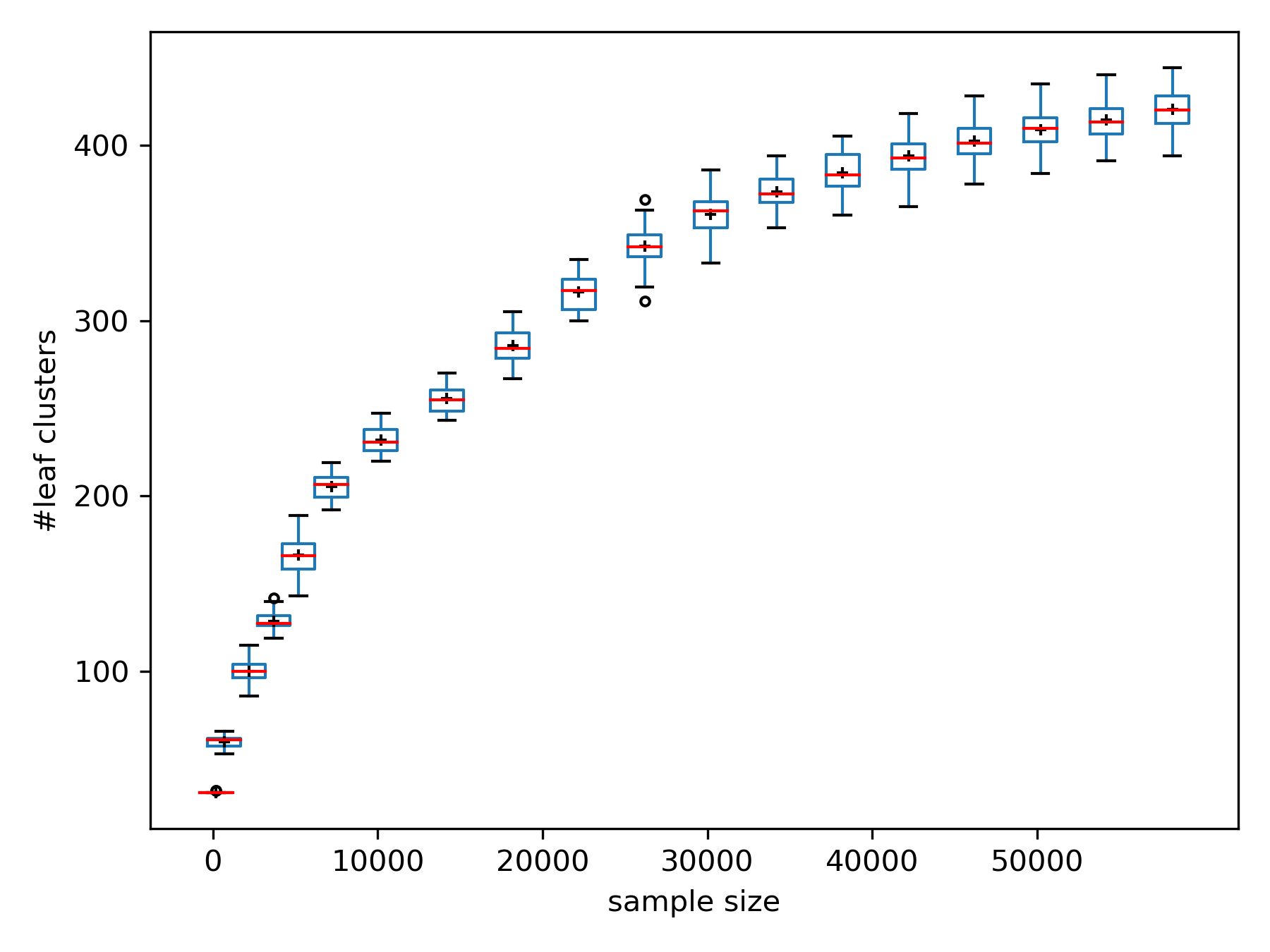}
            \label{fig:srp_num_clusters_rep}        
        } \\
        \subfigure[Depth vs Sample Size]{
            \includegraphics[width=0.32\textheight,height=0.30\textheight]{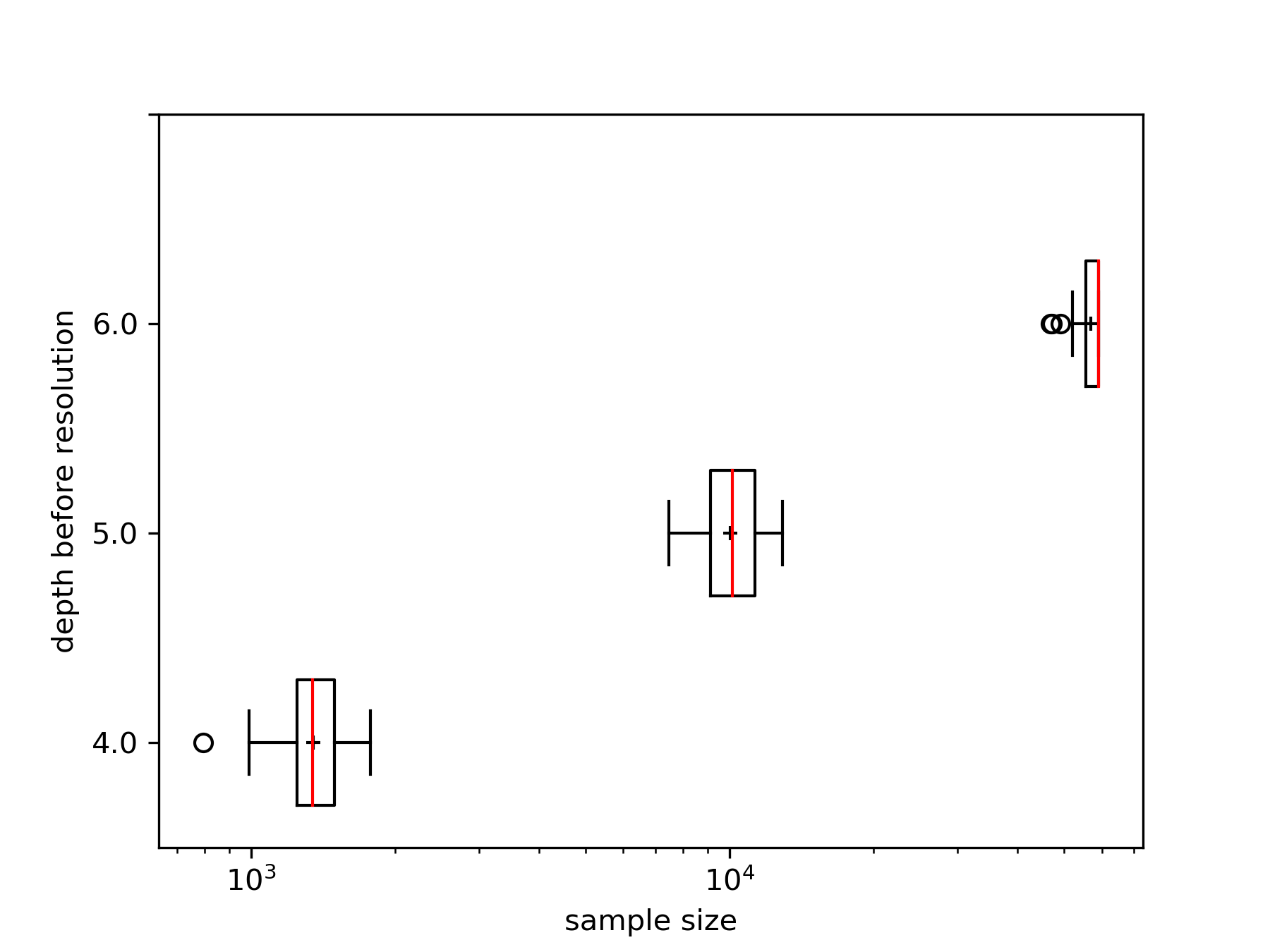}
            \label{fig:srp_depth_rep}        
        }
        \subfigure[Cluster Size]{
            \includegraphics[width=0.32\textheight,height=0.275\textheight]{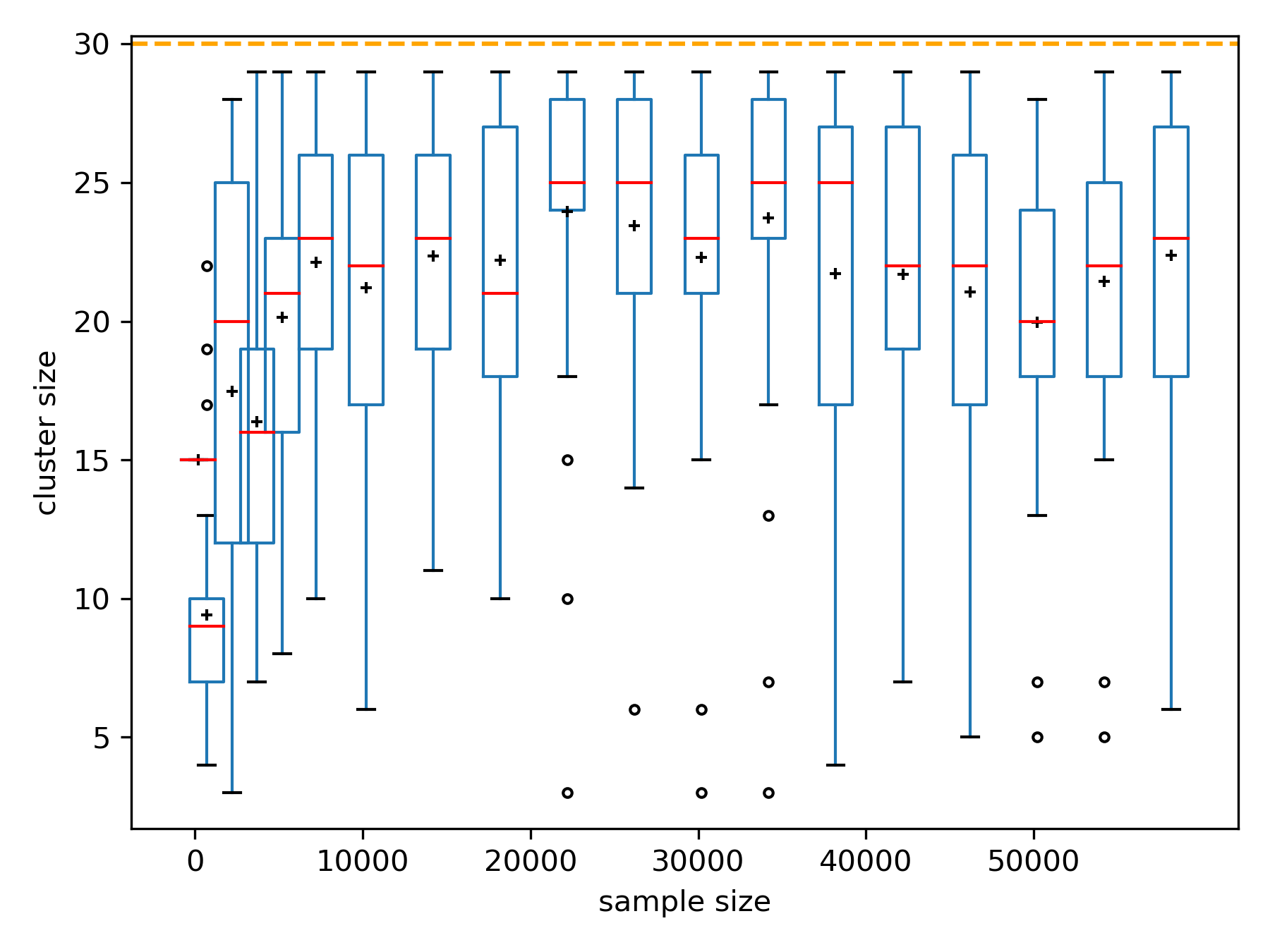}
            \label{fig:srp_numpts_rep}        
        }                 
    \caption{Figure (a) shows the average mean squared error of the approximation (blue) by sample size for the experiment of Swiss Roll intersected with a plane, along with the average maximum MSE (red) at leaf clusters after 30 repeats. The standard error bars are also depicted on each curve. A green horizontal line depicts the threshold level $\epsilon = 0.1$. Figure (b) shows the boxplots of the number of leaf clusters used for the approximation by sample size after 30 repeats. The line bisecting the box represents the median, and the $+$ sign within the box represents the mean number of leaf clusters. Figure (c) illustrates the maximum height of the cluster map. Each boxplot in Figure (c) illustrates, on average, how many new samples it takes before the cluster map needs to resolve an additional depth (based on 30 repeats of the experiment). Figure (d) depicts the boxplots of the number of samples within the leaf cluster where the maximum MSE is observed after each increment (the orange horizontal line shows $M = 30$).}
    \label{fig:srp_mse_num_clusters_rep}
\end{figure}

We use the threshold values $M = 30$ and $\epsilon = 0.5$ for cluster splitting criteria to get a well-resolved manifold. \Cref{fig:srp_train_fit} shows the initial GMRA approximation while the final approximation after streaming in all the data is shown in \cref{fig:srp_stream}. The ground truth is shown in \cref{fig:srp_grouthtruth}. The incrementally updated GMRA approximation visually looks identical to the ground truth. The RMSE error metric value for the difference between ground truth and stream approximation is $0.00135$.
This experiment was also repeated 30 times, and the results are provided in \Cref{fig:srp_mse_num_clusters_rep}, which can be interpreted similarly to \Cref{fig:swiss_mse_num_clusters_rep}.

The obtained experimental results corroborate those observed in \Cref{subsec:case1_swiss_roll}. With the continuous streaming of samples, the algorithm demonstrated its adaptability by dynamically adjusting the number of clusters to better represent the complex manifold's structure, consequently leading to a decrease in the Mean Squared Error (MSE). 

Drawing a comparison to the S-ISOMAP++ algorithm proposed by \cite{mahapatra2017s}, their investigation revolved around approximating intersecting manifolds. However, their approach required samples from both intersecting manifolds for proper functionality. In contrast, our incremental GMRA algorithm only utilized samples from the Swiss roll, autonomously detecting the intersecting plane. This distinctive capability underscores the proposed incremental GMRA's potential in capturing novel manifold behaviors that may emerge in a data stream, a feature not shared by the methodology proposed in \cite{mahapatra2017s}. Nonetheless, it is important to acknowledge that further experiments are essential to substantiate this claim.

\section{Conclusions}

This paper introduces a Non-linear Dimensionality Reduction (NLDR) method designed specifically for streaming data, utilizing the Geometric Multi-Resolution Analysis (GMRA) principles. The primary objective is to assess the incremental GMRA approach's effectiveness compared to the traditional batch GMRA, delving into the intricacies of applying GMRA to streaming scenarios. Challenges tackled include updating the existing cluster map incrementally and determining optimal strategies for updating basis vectors.

The process of updating the cluster map, detailed in \cref{subsec:inc_gmra_alg}, demonstrated simplicity through the application of the Cover Tree method, ensuring consistency with the batch GMRA approach under shared threshold criteria for cluster splitting. While it would suffice to create a cluster map that satisfies the assumptions we make about clusters, achieving an identical match to the cluster map generated by the batch GMRA method provided the added benefit of enabling a direct comparison with ground truth. This allowed for a more comprehensive evaluation of the incremental method and its alignment with established methodologies.

A crucial aspect of this process lies in the choice of stopping criteria for cluster splitting, specifically the Mean Squared Error (MSE) within a given cluster. It is noteworthy that our threshold criteria differ from the one proposed in the Adaptive GMRA version \cite{liao2019adaptive}. The threshold, based on MSE in a given cluster, comes with certain limitations highlighted in Remark 9,  \cite{liao2019adaptive}, particularly when $d \geq 3$. In such cases, to guarantee the rate MSE $\lesssim \left(\frac{\log n}{n}\right)^{\frac{2s}{2s+d-2}}$ proved in Theorem 8, \cite{liao2019adaptive}, the threshold value would have to depend on $s$, a regularity parameter of the Borel probability measure $\mu$, and $d$ is the intrinsic dimension. In our numerical experiments the maximum value of $d$ was 2 therefore, further experimentation is needed for the scenario $d \geq 3$. Following cluster map updates, Brand's approach facilitated efficient incremental updates of basis vectors when handling new data samples.

Transitioning to experimental validation in \Cref{sec:experiments}, simulations explored diverse scenarios, including simple and complex manifolds. The results of performing incremental updates to GMRA confirmed the incremental GMRA's ability to adapt to evolving data, providing comparable approximations even with small initial samples. 

In essence, this work contributes a comprehensive understanding of the incremental GMRA approach's efficacy in NLDR for streaming data. Through theoretical insights, computational efficiencies, and experimental validations, the proposed method showcases adaptability, robustness, and resource optimization in handling dynamic, evolving data streams.

\section{Limitations and Future Work}

Future directions include handling data removal, which would require updating basis vectors, merging clusters, and reassigning cluster members, which are essential for dynamic applications like visual tracking and time-varying manifolds.

The current approach assumes complete data without missing values. Extending Incremental GMRA to handle missing data efficiently remains a challenge, as traditional imputation methods may not integrate well within its structure \cite{balzano2018streaming}.

Finally, the current method assumes a fixed lower dimension, 
$d$ at all scales. However, real-world data often exhibits varying intrinsic dimensions across clusters and scales \cite{camastra2016intrinsic,little2017multiscale}. Future work should develop adaptive strategies to dynamically estimate and adjust $d$ as new data streams in, improving the flexibility of Incremental GMRA.

\appendix
\section{Appendix}
\subsection{Rank \texorpdfstring{$m$}{m} Covariance Matrix Update}
\label{rank_m_update}
Let $\mathbf{X} \in \mathbb{R}^{D\times n}$ be the data set of interest and $\hat{\mathbf{X}} = \begin{bmatrix} \mathbf{X} & \mathbf{C} \end{bmatrix}$ with $\mathbf{C}\in \mathbb{R}^{D\times m}$. The covariance matrix of $\hat{\mathbf{X}}$ is

\begin{align*}
  \cov(\hat{\mathbf{X}}) &= 
  \frac{1}{(n+m-1)}\hat{\mathbf{X}}
  \left(\mathbf{I} - \frac{1}{(n+m)}
  \mathbf{1}_{n+m}\mathbf{1}_{n+m}^\top\right)
  \hat{\mathbf{X}}^\top \\
    &= \frac{1}{(n+m-1)}\begin{bmatrix} \mathbf{X} & \mathbf{C} \end{bmatrix}
    \left(\mathbf{I} - \frac{1}{(n+m)}\mathbf{1}_{n+m}\mathbf{1}_{n+m}^\top\right) \begin{bmatrix} 
    \mathbf{X} & \mathbf{C} \end{bmatrix}^\top 
 \end{align*}
 Multiplying both sides by $(n+m-1)$ for convenience,
 \begin{align*}
    (n+m-1)\cov(\hat{\mathbf{X}}) 
    &=
    \begin{bmatrix} 
    \mathbf{X} & \mathbf{C} 
    \end{bmatrix}
    \begin{bmatrix} 
      \mathbf{X}^\top \\ 
      \mathbf{C}^\top 
    \end{bmatrix} \\ &- 
    \frac{1}{n+m}
    \begin{bmatrix} 
    \mathbf{X} & \mathbf{C} 
    \end{bmatrix}
    \begin{bmatrix}
      \mathbf{1}_n \\
      \mathbf{1}_m
    \end{bmatrix}
    \begin{bmatrix}
      \mathbf{1}_n^\top & 
      \mathbf{1}_m^\top
    \end{bmatrix}
    \begin{bmatrix} 
      \mathbf{X}^\top \\ 
      \mathbf{C}^\top 
    \end{bmatrix} \\
     &=\mathbf{X}\,\mathbf{X}^\top + \mathbf{C}\,\mathbf{C}^\top
     - \frac{1}{n+m}\left( n\,\bar{\mathbf{x}} + m\,\bar{\mathbf{c}} \right) \left( n\,\bar{\mathbf{x}} + m\,\bar{\mathbf{c}} \right)^\top \\
     &= \mathbf{X}\,\mathbf{X}^\top + \mathbf{C}\,\mathbf{C}^\top- 
     \frac{1}{(n+m)}\left(n^2\bar{\mathbf{x}}\,\bar{\mathbf{x}}^\top + n\,m\,\bar{\mathbf{x}}\,\bar{\mathbf{c}}^\top+
     n\,m\,\bar{\mathbf{c}}\,\bar{\mathbf{x}}^\top
     +m^2\bar{\mathbf{c}}\,\bar{\mathbf{c}}^\top\right) 
     \end{align*}
     \begin{align*}
    \frac{(n+m-1)}{(n-1)}\cov(\hat{\mathbf{X}}) &=  \left(\frac{1}{(n-1)}
    \mathbf{X}\,\mathbf{X}^\top -
    \frac{n}{(n-1)}
    \bar{\mathbf{x}}\,\bar{\mathbf{x}}^\top\right) + 
    \frac{(n+m)\,n}{(n-1)(n+m)}
    \bar{\mathbf{x}}\,\bar{\mathbf{x}}^\top \\
      &\phantom{=} \qquad  -\frac{n^2}{(n-1)(n+m)}
    \bar{\mathbf{x}}\,\bar{\mathbf{x}}^\top + 
     \frac{1}{(n-1)}\mathbf{C}\,\mathbf{C}^\top \\ 
    &\phantom{=} \qquad 
    -\frac{1}{(n+m)(n-1)} \left(n\,m\,\bar{\mathbf{x}}\,\bar{\mathbf{c}}^\top
    +n\,m\,\bar{\mathbf{c}}\,\bar{\mathbf{x}}^\top +m^2\bar{\mathbf{c}}\,\bar{\mathbf{c}}^\top\right). 
\end{align*}
With some algebraic manipulation,
\begin{align*}
  \frac{(n+m-1)}{(n-1)}\cov(\hat{\mathbf{X}})
    &=\cov(\mathbf{X})+\frac{n\,m}{(n-1)(n+m)}\left(\bar{\mathbf{x}}\,\bar{\mathbf{x}}^\top -\bar{\mathbf{x}}\,\bar{\mathbf{c}}^\top-\bar{\mathbf{c}}\,\bar{\mathbf{x}}^\top\right)\\
    &\phantom{=}\qquad
    +\frac{1}{(n-1)}\mathbf{C}\,\mathbf{C}^\top-
    \frac{m^2}{(n+m)(n-1)}\bar{\mathbf{c}}\,\bar{\mathbf{c}}^\top \\
    &=\cov(\mathbf{X})+\frac{n\,m}{(n-1)(n+m)}\left(\bar{\mathbf{x}}\,\bar{\mathbf{x}}^\top -\bar{\mathbf{x}}\,\bar{\mathbf{c}}^\top-\bar{\mathbf{c}}\,\bar{\mathbf{x}}^\top+
    \bar{\mathbf{c}}\,\bar{\mathbf{c}}^\top\right)\\
    &\phantom{=}\qquad 
     -\frac{n\,m}{(n-1)(n+m)} \bar{\mathbf{c}}\,\bar{\mathbf{c}}^\top
   +\frac{1}{(n-1)}\mathbf{C}\,\mathbf{C}^\top\\
   &\phantom{=}\qquad
    -\frac{m^2}{(n+m)(n-1)}\bar{\mathbf{c}}\bar{\mathbf{c}}^\top \\
  &=\cov(\mathbf{X})+\frac{n\,m}{(n-1)(n+m)}(\bar{\mathbf{x}}-\bar{\mathbf{c}})(\bar{\mathbf{x}}-\bar{\mathbf{c}})^\top\\
    &\phantom{=}\qquad +
    \frac{1}{(n-1)}
    \mathbf{C}\,\mathbf{C}^\top-
    \frac{m(n+m)}{(n-1)(n+m)}
    \bar{\mathbf{c}}\,\bar{\mathbf{c}}^\top \\
 &=\cov(\mathbf{X})+\frac{n\,m}{(n-1)(n+m)}(\bar{\mathbf{x}}-\bar{\mathbf{c}})(\bar{\mathbf{x}}-\bar{\mathbf{c}})^\top
+   \frac{(m-1)}{(n-1)} \cov{(\mathbf{C})}
\end{align*}
Multiplying both sides by $\displaystyle \frac{(n-1)}{(n+m-1)}$ yields the desired result,
\begin{align*}
    \cov(\hat{\mathbf{X}})&=\frac{(n-1)}{(n+m-1)}\cov(\mathbf{X})+\frac{nm}{(n+m-1)(n+m)}(\bar{\mathbf{x}}-\bar{\mathbf{c}})(\bar{\mathbf{x}}-\bar{\mathbf{c}})^\top\\
    &\phantom{=} \qquad
    +\frac{(m-1)}{(n+m-1)}\cov(\mathbf{C}) 
\end{align*}
where 
\begin{align*}
    \cov(\mathbf{C}) &= \frac{1}{(m-1)}\sum_{i = n+1}^{n+m}({\mathbf{x}_i-\bar{\mathbf{c}}})({\mathbf{x}_i-\bar{\mathbf{c}}})^\top \\
    \mathbf{X} &=[\mathbf{x}_1,\, \mathbf{x}_2,\cdots,\mathbf{x}_n] \\
    \hat{\mathbf{X}} &=[\mathbf{x}_1,\, \mathbf{x}_2,\cdots,\mathbf{x}_{n+m}]
\end{align*}
\newpage
\subsection{Proof of Lemma 1}
\label{appendix:lemma1_proof}
\begin{proof}
Let $\mathbf{C} = \mathbf{C}^\top \in\mathbb{R}^{D\times D}$. For convenience, with a slight abuse of notation, we will ignore the multiplication by the constants $a$ and $b$ and assume that $\mathbf{C}$ and $\mathbf{B}\,\mathbf{B}^\top$ have been already multiplied by them. 

We wish to bound
\begin{align}
\left\|
\sigma_i \left( \mathbf{C} + \mathbf{B}\,\mathbf{B}^\top \right) - 
\sigma_i \left(\mathbf{C}_d + \mathbf{B}\,\mathbf{B}^\top \right)  
\right\|
\label{eqn:goal1}
\end{align}
where $d<D$.

The Courant--Fisher Theorem (1905),Theorem 8.1.5 \cite{MR1417720} gives us useful inequalities to compare the singular values of \cref{eqn:Rd1,eqn:Rd2}.
\begin{align}
\label{eqn:CF_1}
\sigma_i(\mathbf{C}) + \sigma_D(\mathbf{B}\,\mathbf{B}^\top)
&\le \sigma_i(\mathbf{C} + \mathbf{B}\,\mathbf{B}^\top)
\le \sigma_i(\mathbf{C}) + \sigma_1(\mathbf{B}\,\mathbf{B}^\top) \\
\label{eqn:CF_2}
\sigma_i(\mathbf{C}_d) + \sigma_D(\mathbf{B}\,\mathbf{B}^\top)
&\le \sigma_i(\mathbf{C}_d + \mathbf{B}\,\mathbf{B}^\top)
\le \sigma_i(\mathbf{C}_d) + \sigma_1(\mathbf{B}\,\mathbf{B}^\top) \\
\label{eqn:CF_3}
\sigma_i(\mathbf{B}\,\mathbf{B}^\top) + \sigma_D(\mathbf{C})
&\le \sigma_i(\mathbf{C} + \mathbf{B}\,\mathbf{B}^\top)
\le \sigma_i(\mathbf{B}\,\mathbf{B}^\top) + \sigma_1(\mathbf{C}) \\
\label{eqn:CF_4}
\sigma_i(\mathbf{B}\,\mathbf{B}^\top) + \sigma_D(\mathbf{C}_d)
&\le \sigma_i(\mathbf{C}_d + \mathbf{B}\,\mathbf{B}^\top)
\le \sigma_i(\mathbf{B}\,\mathbf{B}^\top) + \sigma_1(\mathbf{C}_d). 
\end{align}
We can simplify these inequalities slightly using
$\sigma_i(\mathbf{B}\,\mathbf{B}^\top) = \sigma_i^2(\mathbf{B})$, $\sigma_D(\mathbf{C}_d) = 0$ and $\sigma_1(\mathbf{C}) = \sigma_1(\mathbf{C}_d)$. Although there are four variations of inequalities, \cref{eqn:CF_1,eqn:CF_2} are the ones we need to consider. 
\begin{align}
\label{eqn:d8}
\sigma_i(\mathbf{C}) + \sigma_D^2(\mathbf{B})
&\le \sigma_i(\mathbf{C} + \mathbf{B}\,\mathbf{B}^\top)
\le \sigma_i(\mathbf{C}) + \sigma_1^2(\mathbf{B}) \\
\label{eqn:d9}
\sigma_i(\mathbf{C}_d) + \sigma_D^2(\mathbf{B})
&\le \sigma_i(\mathbf{C}_d + \mathbf{B}\,\mathbf{B}^\top)
\le \sigma_i(\mathbf{C}_d) + \sigma_1^2(\mathbf{B}) 
\end{align}
\Cref{eqn:goal1} is equivalent to maximizing
\begin{align}
\left| \sigma_i(\mathbf{C} + \mathbf{B}\,\mathbf{B}^\top) - \sigma_i(\mathbf{C}_d + \mathbf{B}\,\mathbf{B}^\top)\right|
\label{eqn:goal2}
\end{align}
for some (unknown) index $i$. 
First, consider the case when 
\begin{align}
\sigma_i(\mathbf{C} + \mathbf{B}\,\mathbf{B}^\top) > \sigma_i(\mathbf{C}_d + \mathbf{B}\,\mathbf{B}^\top).
\end{align}
To maximize the difference, we want $\sigma_i(\mathbf{C} + \mathbf{B}\,\mathbf{B}^\top)$ to be as large as possible (upper bound), and 
$\sigma_i(\mathbf{C}_d + \mathbf{B}\,\mathbf{B}^\top)$ to be as small as possible (lower bound). 

Upper bound in \cref{eqn:d8} and lower bound in \cref{eqn:d9} gives the following bound
\begin{align}
\left| \sigma_i(\mathbf{C} + \mathbf{B}\,\mathbf{B}^\top) - \sigma_i(\mathbf{C}_d + \mathbf{B}\,\mathbf{B}^\top)\right| &\le 
\sigma_i(\mathbf{C}) + \sigma_1^2(\mathbf{B}) 
- \sigma_i(\mathbf{C}_d) - \sigma_D^2(\mathbf{B}) 
\end{align}
Notice that $\sigma_i(\mathbf{C}) = \sigma_i(\mathbf{C}_d)$ for $i \le d$ and $\sigma_i(\mathbf{C}) - 
\sigma_i(\mathbf{C}_d) \leq \sigma_{d+1}(\mathbf{C})$ for $i>d$. 

Furthermore, if $n<D$ then $\sigma_D^2(\mathbf{B}) = 0$. Therefore, we get
\begin{align}
\label{eqn:bound1}
\left| \sigma_i(\mathbf{C} + \mathbf{B}\,\mathbf{B}^\top) - \sigma_i(\mathbf{C}_d + \mathbf{B}\,\mathbf{B}^\top)\right| &\le 
\sigma_{d+1}(\mathbf{C}) + \sigma_1^2(\mathbf{B}) 
\end{align}
In the second case 
\begin{align}
\sigma_i(\mathbf{C} + \mathbf{B}\,\mathbf{B}^\top) < \sigma_i(\mathbf{C}_d + \mathbf{B}\,\mathbf{B}^\top).
\end{align}
we want $\sigma_i(\mathbf{C} + B\,B^\top)$ to be as small as possible (lower bound), and 
$\sigma_i(\mathbf{C}_d + \mathbf{B}\,\mathbf{B}^\top)$ to be as large as possible (lower bound).

Lower bound in \cref{eqn:d8} and upper bound in \cref{eqn:d9} gives the following bound
\begin{align}
\label{eqn:bound2}
\left| \sigma_i(\mathbf{C} + \mathbf{B}\,\mathbf{B}^\top) - \sigma_i(\mathbf{C}_d + \mathbf{B}\,\mathbf{B}^\top)\right| & =  
\left| \sigma_i(\mathbf{C}) + \sigma_D^2(\mathbf{B}) 
- \sigma_i(\mathbf{C}_d) - \sigma_1^2(\mathbf{B}) \right | \\
&\le \sigma_1^2(\mathbf{B}) - \sigma_{d+1}(\mathbf{C}) 
\end{align}
\Cref{eqn:bound1,eqn:bound2} imply that 
\begin{align}
\left| \sigma_i(\mathbf{C} + \mathbf{B}\,\mathbf{B}^\top) - \sigma_i(\mathbf{C}_d + \mathbf{B}\,\mathbf{B}^\top)\right|
&\le \sigma_{d+1}(\mathbf{C}) +\sigma_1^2(\mathbf{B})
\end{align}
\end{proof}

\subsection{Proof of Lemma 2}
\begin{proof}
Let $ \mathbf{A}_1 := a\,\mathbf{C} + b\,\mathbf{B}\,\mathbf{B}^\top$,
$ \mathbf{A}_2 := a\,\mathbf{C}_d + b\,\mathbf{B}\,\mathbf{B}^\top$,  $\Sigma = diag(\lambda_1,\ldots, \lambda_D)$ where $a = \frac{n-1}{(n+m-1)}$, $b = \frac{1}{(n+m-1)}$.
\begin{align*}
\mathbf{A}_1 &= \mathbf{A}_2 + a\,\sum_{i= d+1}^D\lambda_i\,u_i\,u_i^\top \\
 &= a\,\mathbf{C}_d + b\,\mathbf{B}\,\mathbf{B}^\top + a\,\sum_{i= d+1}^D\lambda_i\,u_i\,u_i^\top 
\end{align*}

Therefore, using Theorem 3 \cite{zwald2005convergence} we can write
\begin{align}
    \label{proof:angle_proof_step1}
    \|\tilde{\mathbf{R}}_d\,\tilde{\mathbf{R}}_d^\top - 
    {\mathbf{R}}_d\,{\mathbf{R}}_d^\top\|_2 \leq 
    \frac{\|a\,\sum_{i= d+1}^D\lambda_i\,u_i\,u_i^\top\|_2}
    {\delta_d}
\end{align}
where $\delta_d = 0.5\left(\sigma_d(a\,\mathbf{C}_d+b\,\mathbf{B}\,\mathbf{B}^\top)
- \sigma_{d+1}(a\,\mathbf{C}_d+b\,\mathbf{B}\,\mathbf{B}^\top)
\right)$.
We want this eigengap to be large. Therefore, we need 
$\sigma_d(a\,\mathbf{C}_d+b\,\mathbf{B}\,\mathbf{B}^\top)$ large as much as possible and
$\sigma_{d+1}(a\,\mathbf{C}_d+b\,\mathbf{B}\,\mathbf{B}^\top)$ smaller as much as possible.

For convenience, with a slight abuse of notation, we will ignore the multiplication by the constants $a$ and $b$ and assume that $\mathbf{C}_d$ and $\mathbf{B}\,\mathbf{B}^\top$ have been already multiplied by them.  
Using \cref{eqn:CF_2,eqn:CF_4} with setting $i=d$ and $i=d+1$ we get the following inequalities.
\begin{align}
    \label{eqn:thm_p1}
    \sigma_d(\mathbf{C}_d) + \sigma_D(\mathbf{B}\,\mathbf{B}^\top)
    &\le \sigma_d(\mathbf{C}_d + \mathbf{B}\,\mathbf{B}^\top)
    \le \sigma_d(\mathbf{C}_d) + \sigma_1(\mathbf{B}\,\mathbf{B}^\top) \\
    \label{eqn:thm_p2}
    \sigma_d(\mathbf{B}\,\mathbf{B}^\top) + \sigma_D(\mathbf{C}_d)
    &\le \sigma_d(\mathbf{C}_d + \mathbf{B}\,\mathbf{B}^\top)
    \le \sigma_d(\mathbf{B}\,\mathbf{B}^\top) + \sigma_1(\mathbf{C}_d). \\
    \label{eqn:thm_p3}
    \sigma_{d+1}(\mathbf{C}_d) + \sigma_D(\mathbf{B}\,\mathbf{B}^\top)
    &\le \sigma_{d+1}(\mathbf{C}_d + \mathbf{B}\,\mathbf{B}^\top)
    \le \sigma_{d+1}(\mathbf{C}_d) + \sigma_1(\mathbf{B}\,\mathbf{B}^\top) \\
    \label{eqn:thm_p4}
    \sigma_{d+1}(\mathbf{B}\,\mathbf{B}^\top) + \sigma_D(\mathbf{C}_d)
    &\le \sigma_{d+1}(\mathbf{C}_d + \mathbf{B}\,\mathbf{B}^\top)
    \le \sigma_{d+1}(\mathbf{B}\,\mathbf{B}^\top) + \sigma_1(\mathbf{C}_d). 
\end{align}
We want to make the upper bound depend on the largest singular value of the $\mathbf{B}$. Therefore, we consider the largest value from \cref{eqn:thm_p1} and the smallest value from \cref{eqn:thm_p3}. Notice that 
$\sigma_{d+1}(\mathbf{C}_d)$ and $\sigma_D(\mathbf{B}\,\mathbf{B}^\top)$ are both zero. So we get,
\begin{align*}
   \sigma_d(\mathbf{C}_d+\mathbf{B}\,\mathbf{B}^\top)
- \sigma_{d+1}(\mathbf{C}_d+\mathbf{B}\,\mathbf{B}^\top) &=
\sigma_d(\mathbf{C}_d) + \sigma_1^2(\mathbf{B})
\end{align*}
Therefore we can rewrite \Cref{proof:angle_proof_step1},
\begin{align}
    \|\tilde{\mathbf{R}}_d\,\tilde{\mathbf{R}}_d^\top - 
    {\mathbf{R}}_d\,{\mathbf{R}}_d^\top\|_2 &\leq 
    \frac{\|a\,\,\sum_{i= d+1}^D\lambda_i\,u_i\,u_i^\top\|_2}
    {b\,\delta_d}
    \leq 
    \frac{2\,a\,\sigma_{d+1}(\mathbf{C})}{a\,\sigma_d(\mathbf{C}_d) + b\,\sigma_1^2(\mathbf{B}))}
\end{align}
\end{proof}

 \bibliographystyle{unsrtnat} 
 \bibliography{references}

\end{document}